\documentclass[lettersize,journal]{IEEEtran}
\usepackage{amsmath,amsfonts}
\usepackage{amssymb}
\usepackage{amsthm}
\usepackage{cases}
\usepackage{algorithmic}
\usepackage{algorithm}
\usepackage{array}
\usepackage{booktabs}
\usepackage{multirow}
\usepackage{subcaption}
\usepackage{textcomp}
\usepackage{stfloats}
\usepackage{url}
\usepackage{verbatim}
\usepackage{graphicx}
\usepackage{cite}
\usepackage{pifont}
\usepackage[accsupp]{axessibility}
\usepackage[table]{xcolor}
\allowdisplaybreaks

\DeclareMathOperator{\argmin}{arg\,min}

\newtheorem{assumption}{Assumption}[section]
\newtheorem{theorem}{Theorem}[section]
\newtheorem{lemma}[theorem]{Lemma}

\newtheorem{corollary}{Corollary}[theorem]
\newtheorem{remark}{Remark}[theorem]

\makeatletter
\def\mathcolor#1#{\@mathcolor{#1}}
\def\@mathcolor#1#2#3{%
  \protect\leavevmode
  \begingroup
    \color#1{#2}#3%
  \endgroup
}
\makeatother

\begin{document}

\title{E-3SFC: Communication-Efficient Federated Learning with Double-way Features Synthesizing}

\author{Yuhao Zhou,~\IEEEmembership{Student Member,~IEEE}, Yuxin Tian,~\IEEEmembership{Student Member,~IEEE}, Mingjia Shi,~\IEEEmembership{Student Member,~IEEE}, Yuanxi Li, Yanan Sun,~\IEEEmembership{Senior Member,~IEEE}, Qing Ye, Jiancheng Lv,~\IEEEmembership{Senior Member,~IEEE}
        % <-this % stops a space
\thanks{Yuhao Zhou (e-mail: sooptq@gmail.com), Yuxin Tian (e-mail: cs.yuxintian@outlook.com), Mingjia Shi (e-mail: 3101ihs@gmail.com), Yanan Sun (e-mail: ysun@scu.edu.cn), Qing Ye (e-mail: yeqing@scu.edu.cn) and Jiancheng Lv (e-mali: lvjiancheng@scu.edu.cn) are with the College of Computer Science, Sichuan University, Chengdu 610065, P. R. China and Engineering Research Center of Machine Learning and Industry Intelligence, Ministry of Education, Chengdu 610065, P. R. China. Yuanxi Li (e-mail: yuanxi3@illinois.edu) is with the University of Illinois at Urbana-Champaign}% <-this % stops a space
\thanks{Yuhao Zhou and Yuxin Tian have equal contributions to this manuscript.}
\thanks{Qing Ye is the corresponding author.}
\thanks{© 2025 IEEE.  Personal use of this material is permitted.  Permission from IEEE must be obtained for all other uses, in any current or future media, including reprinting/republishing this material for advertising or promotional purposes, creating new collective works, for resale or redistribution to servers or lists, or reuse of any copyrighted component of this work in other works.}}
% \thanks{Manuscript received February 11, 2024; accepted Febrary 3, 2025.}}

\markboth{Journal of \LaTeX\ Class Files,~Vol.~14, No.~8, August~2021}%
{Zhou \MakeLowercase{\textit{et al.}}: A Sample Article Using IEEEtran.cls for IEEE Journals}


\maketitle

\begin{abstract}
The exponential growth in model sizes has significantly increased the communication burden in Federated Learning (FL).
Existing methods to alleviate this burden by transmitting compressed gradients often face high compression errors, which slow down the model’s convergence.
To simultaneously achieve high compression effectiveness and lower compression errors, we study the gradient compression problem from a novel perspective.
Specifically, we propose a systematical algorithm termed Extended Single-Step Synthetic Features Compressing (E-3SFC), which consists of three sub-components, \textit{i.e.}, the Single-Step Synthetic Features Compressor (3SFC), a double-way compression algorithm, and a communication budget scheduler.
First, we regard the process of gradient computation of a model as decompressing gradients from corresponding inputs, while the inverse process is considered as compressing the gradients.
Based on this, we introduce a novel gradient compression method termed 3SFC, which utilizes the model itself as a decompressor, leveraging training priors such as model weights and objective functions.
3SFC compresses raw gradients into tiny synthetic features in a single-step simulation, incorporating error feedback to minimize overall compression errors.
To further reduce communication overhead, 3SFC is extended to E-3SFC, allowing double-way compression and dynamic communication budget scheduling.
Our theoretical analysis under both strongly convex and non-convex conditions demonstrates that 3SFC achieves linear and sub-linear convergence rates with aggregation noise.
Extensive experiments across six datasets and six models reveal that 3SFC outperforms state-of-the-art methods by up to 13.4\% while reducing communication costs by 111.6 times.
These findings suggest that 3SFC can significantly enhance communication efficiency in FL without compromising model performance.
\end{abstract}

\begin{IEEEkeywords}
Federated Learning, Communication-Efficient, Gradient Compression.
\end{IEEEkeywords}

\section{Introduction}
\IEEEPARstart{F}{ederated} learning~\cite{mcmahan2017communication} (FL) aims to tackle the isolated data island problem with privacy guarantees~\cite{li2020review}, achieving promising results~\cite{dean2012large,ben2019demystifying,shi2023prior,zhou2024defta}.
Like its counterparts in the realm of distributed computing, the communication costs between the server and clients are crucial for FL~\cite{sattler2019robust,zhou2021communication}.
Nevertheless, the limited network bandwidth~\cite{kairouz2021advances} and explosive growth of model size~\cite{brown2020language,DBLP:conf/icmcs/TianY0LR0024,DBLP:journals/tsmc/TianLYWPL24,fedus2021switch} not only decrease the training efficiency of FL but also hinder the deployment and delivery of FL at the scale~\cite{bonawitz2019towards}.

\begin{figure}[tb]
    \centering
    \includegraphics[width=0.45\textwidth]{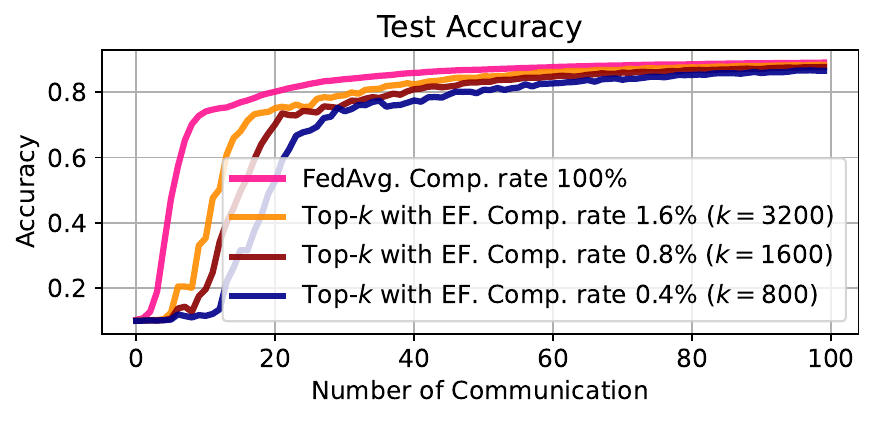}
    \caption{\textbf{Information Compression Rate v.s. Model Convergence}: The rate of convergence reduces as the compression rate decreases. The evaluated MLP model is trained on non-iid MNIST with 20 FL clients.}
    \label{fig:pre-degraded-convergence-rate}
\end{figure}

To cut down the communication cost, existing studies compress the transmitted information, \textit{i.e.,} the gradients of model or model parameters, with diverse compression techniques, such as sparsification~\cite{strom2015scalable,lin2017deep,wangni2018gradient}, quantification~\cite{alistarh2017qsgd,bernstein2018signsgd,karimireddy2019error} and others~\cite{gou2021knowledge,goetz2020federated,li2021communication,hu2022fedsynth}.
However, these methods inevitably make a trade-off between model convergence rate and communication overhead.
More specifically, to achieve lower communication overhead, a higher information compression ratio is usually required, which in turn leads to a limited model convergence rate and performance, as shown in Figure~\ref{fig:pre-degraded-convergence-rate}.
This is consistent with~\cite{aji2017sparse}, as the compression error upper bound of the top-$k$ sparsifier's $||x-topk(x)||_1 \leq (n-k)||x||_1/n$ is positively correlated with the compression rate hyperparameter $k$.
In addition to such a trade-off, the static round-based communication budget constraint in previous studies~\cite{lin2017deep, tang2024z} inevitably suffers from the adverse impact of the accumulation of the compressing error, as the compression is not lossless.
Therefore, it is highly expected to propose a communication-efficient federated learning approach that achieves a high model convergence rate, a low information compression error, and a high training efficiency.

\begin{figure}
     \centering
     \begin{subfigure}[tb]{0.5\textwidth}
         \centering
         \includegraphics[width=\textwidth]{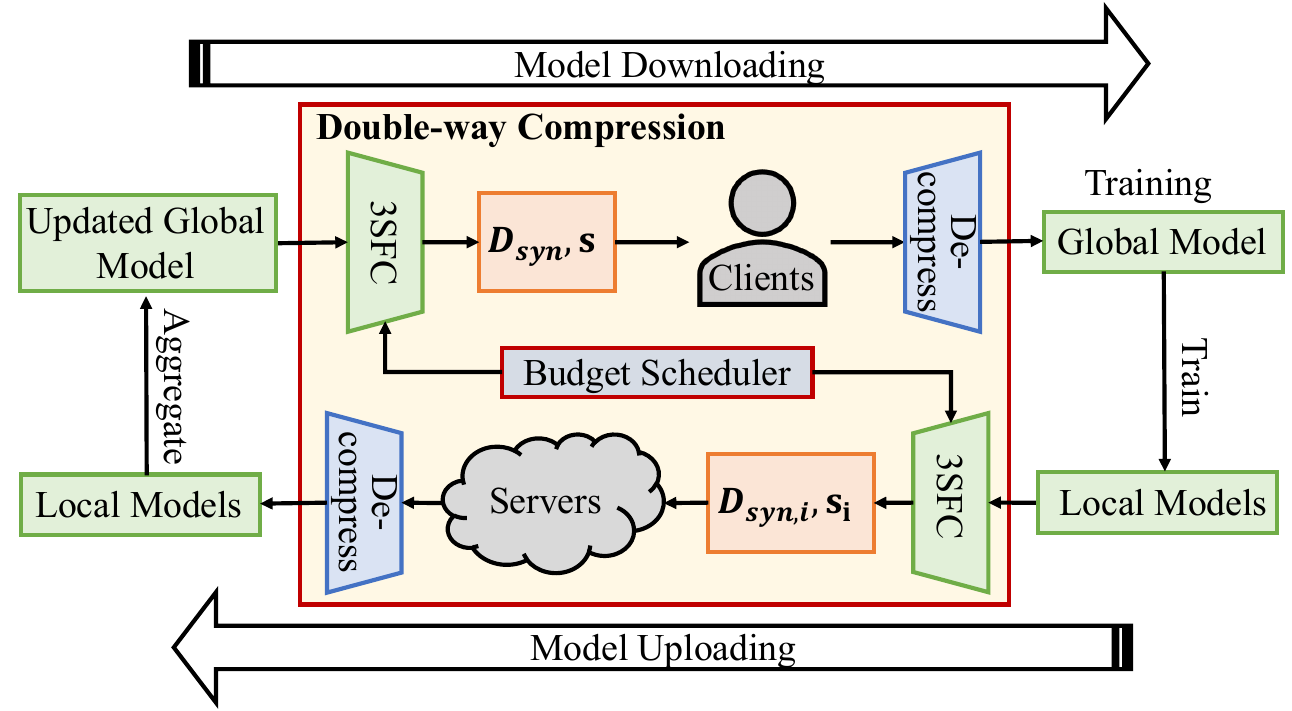}
         \caption{
         Overview of the proposed E-3SFC: It reduces the communication costs in two phases (uploading and downloading) and introduces budget schedulers under the relaxed communication constraint to enhance system flexibility and efficiency.
         }
         \label{fig:e-3sfc-arch}
     \end{subfigure}
     \\
     \begin{subfigure}[tb]{0.5\textwidth}
         \centering
         \includegraphics[width=\textwidth]{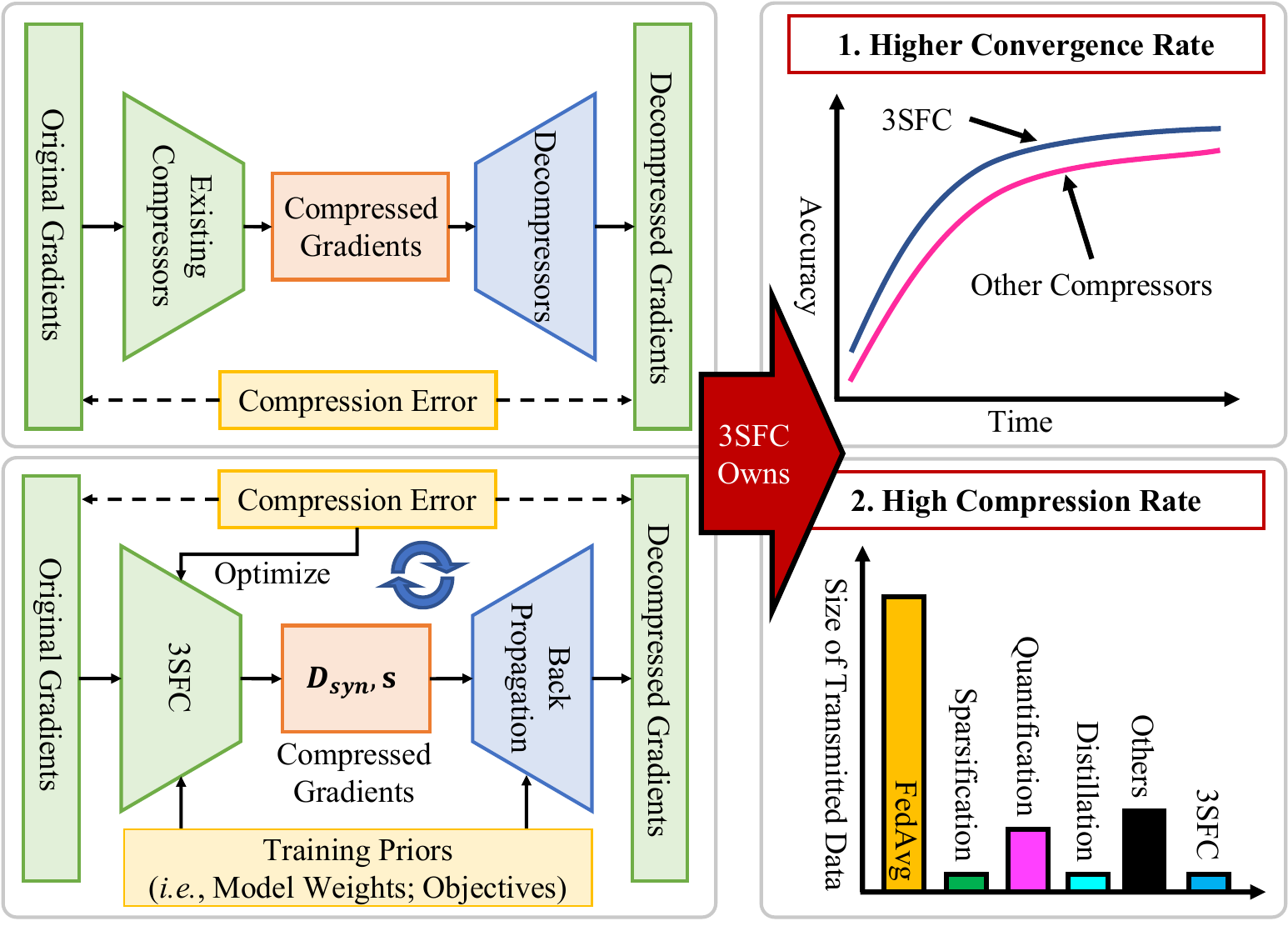}
         \caption{The proposed gradient compressor, \textit{i.e,.} 3SFC, achieves a higher convergence rate while maintaining a high compression rate. It injects training priors into the compression process and utilizes the model itself for back-propagation to optimize compression errors, which provides a novel view of gradient compression.}
         \label{fig:motivation}
     \end{subfigure}
\end{figure}

Different from existing studies that only seek various compression techniques~\cite{sattler2019robust, bernstein2018signsgd, lin2017deep}, we intend to address this challenge by introducing a novel information compressor and a systematic communication workflow optimization.
On the one hand, for the compressor, we consider such a compression-decompression problem from a novel perspective: \textit{the input features can be the compressed gradients and the corresponding network backward pass can be its decompression process.}
More specifically, inspired by~\cite{bottou2018optimization, zhu2019deep}, we deem the gradient computation as a function concerning the training priors, namely the model parameters, the objective function, and the model inputs.
By freezing the training priors, the gradients can be compressed into the synthetic features via backpropagation, which are much smaller than the corresponding gradients.
Meanwhile, by freezing the synthetic features and unfreezing the training priors, the backward pass of such synthetic features can be the corresponding decompression process.
On the other hand, from a systematic perspective, only compressing the transmitted information is not enough for high communication effectiveness and training efficiency and therefore we would like to optimize the workflow of communication-efficient FL.
First, unlike existing methods that only compress the uploaded information~\cite{sattler2019robust, tang2024z}, the communication costs will be further cut by compressing the uploaded and downloaded information.
Then, by replacing the static round-based communication budget constraint with the cumulative-based communication budget, one could adopt a budget scheduler to modulate the budget of each round to alleviate the adverse impact of compression error cumulation.

Based on the above discussions, in this work, we propose a novel communication-efficient FL approach, dubbed Extended Single-Step Synthetic Features Compressing Algorithm (E-3SFC), which not only compresses the information with an insightful perspective and reduces the communication costs of FL as much as possible. 
To be specific, we first introduce a novel \textit{gradient compressor}, namely, Single-Step Synthetic Features Compressor (3SFC), which achieves lower compression errors and thereby boosts the convergence rate, as Figure~\ref{fig:motivation} shows.
In detail, with the determined training priors, 3SFC constructs tiny learnable features as the compressed gradients and adopts error feedback~\cite{stich2018sparsified} (EF) to further reduce the accumulated compression errors.
The advantages of such a compressor can be two-fold.
On the one hand, the compression error is related to the size of synthetic features, and such compression could be lossless if synthetic features have the same shape of raw inputs~\cite{izzo2023theoretical}.
On the other hand, the size of synthetic features is more communication-efficient than original gradients.
For instance, synthetic features of ViT-L/16~\cite{dosovitskiy2020image} ($M \times 224 \times 224 \times 3$ parameters, $M$ is batch size) are much smaller than its corresponding gradients ($304,326,632$ parameters).
Clearly, the compression could not be ideal~\cite{alistarh2018convergence} and lossless, and thus we provide comprehensive theoretical analyses to show that the proposed 3SFC is ensured to have linear and sub-linear convergence rates under strongly convex and non-convex cases with aggregation noise, respectively.
Second, to reach higher training efficiency and communication effectiveness, we optimize the workflow of communication-efficient FL.
The proposed E-3SFC performs \textit{double-way compression} by applying the proposed 3SFC to the entire communication phases of FL to reduce the communication overhead as much as possible.
E-3SFC adopts a budget scheduler to endow a dynamic budget for each round, which assigns more communication budget for the early phase of the FL training to alleviate the negative effects of compression error accumulation.
To evaluate the effectiveness of the proposed algorithm, we conduct extensive experiments with six models on six challenging datasets across Computer Vision (CV) and Natural Language Processing (NLP). 

To sum up, our contributions can be summarized as follows:

\begin{enumerate}
    \item We propose a novel gradient compression algorithm, E-3SFC, to reduce the communication overhead of FL. First, we provide a novel perspective on gradient compression in FL, regarding the model itself as a gradient decompressor. Based on this, we propose a novel gradient compressor, termed 3SFC, to achieve a higher compression rate and lower compression errors. Then we optimize the workflow of the communication-efficient FL by compressing the information of all communication phases to reduce communication cost further and assigning more communication budget to the early FL training stage to reduce the accumulation compression error. To the best of our knowledge, this paper is one of the first attempts to deem the model itself as a gradient decompressor.
    \item Theoretical convergence analyses are provided, showing that the proposed compressor 3SFC has a $\mathcal{O}(\frac{1}{R})$/$\mathcal{O}(\frac{1}{R^2})$ speedup with/without aggregation noise under the strongly convex case, and a $\mathcal{O}(\frac{1}{R^{1/2}})$/$\mathcal{O}(\frac{1}{R^{2/3}})$ convergence rate with/without aggregation noise under the non-convex case.
    \item Extensive experiments are conducted on six datasets and six models, demonstrating that E-3SFC outperforms other methods by up to 13.4\% with 111.6$\times$ less communication costs. Other experimental analyses further validate the effectiveness of E-3SFC's design choices\footnote{https://github.com/Soptq/e-3sfc}.
\end{enumerate}

\begin{figure}[tb]
    \centering
    \includegraphics[width=0.45\textwidth]{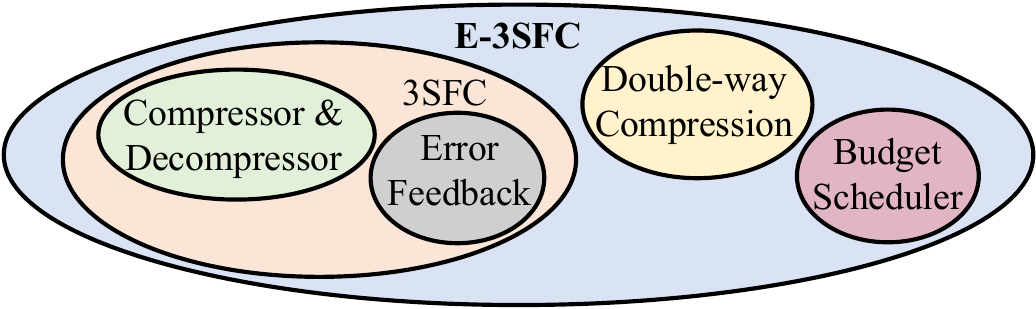}
    \caption{Relationship between E-3SFC and 3SFC}
    \label{fig:relationship}
\end{figure}

Notably, there is a short conference version of this paper appeared in~\cite{zhou2023communication}, which does not optimize the workflow of communication-efficient FL and lacks comprehensive theoretical convergence analyses.
The relationship between E-3SFC and 3SFC is shown in Figure~\ref{fig:relationship}.
Consequently, this paper extends it in the following ways:
1) We propose E-3SFC that optimizes the workflow of communication-efficient FL to further cut the communication overhead whilst 3SFC does not;
2) We provide comprehensive theoretical analyses to ensure the convergence of the proposed 3SFC under mild assumptions;
3) We provide extensive empirical evaluations on CV and NLP benchmarks with a larger number of FL clients to further demonstrate the effectiveness of the proposed E-3SFC.

The rest of this paper is organized as follows. 
Related work is illustrated in Section~\ref{sec:related-work}. 
Section~\ref{sec:problem-formulation} formulates the problem we tackled. 
In Section~\ref{sec:main-method}, details of the proposed E-3SFC are presented. 
Then convergence analyses are described in Section~\ref{sec:theoretical-analysis}, and experiments are documented in Section~\ref{sec:experiments} and Section~\ref{sec:experimental-analysis}. 
Finally, Section~\ref{sec:conclusion} concludes the paper.
\begin{figure}[t]
    \centering
    \includegraphics[width=0.45\textwidth]{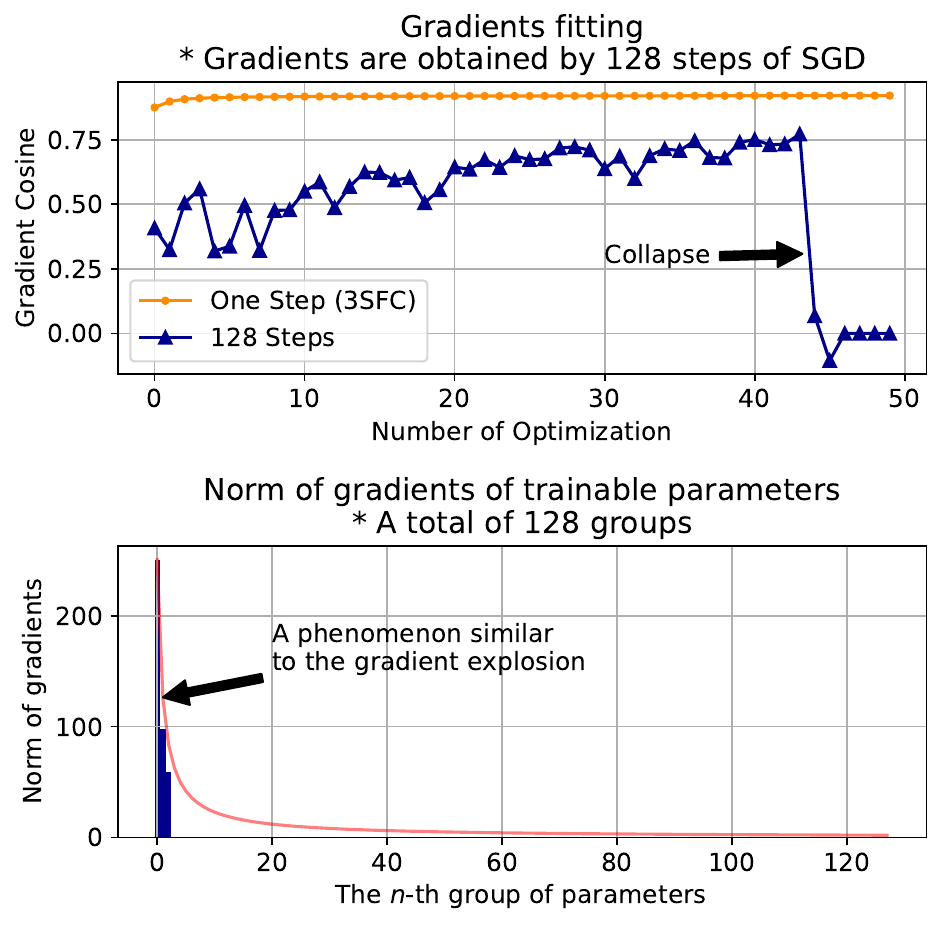}
    \caption{\textbf{Above}: When fitting gradients obtained by 128 steps of SGD using 128 steps of simulation~\protect\cite{goetz2020federated}, it collapsed. Meanwhile, E-3SFC requires only one step of simulation, occupying less computation and storage, but achieves significantly better results. \textbf{Below}: Before the collapse, the gradients of its trainable parameters exhibit a phenomenon similar to the gradient explosion, where the magnitude of gradients increases as they back-propagate from the 128-th to the first group of parameters. This could be a possible reason for the collapse.}
    \label{fig:pre-exp-one}
\end{figure}

\section{Related Work}
\label{sec:related-work}
Works related to communication-efficient FL~\cite{chen2021communication} can be mainly categorized into the following four topics: sparsification, quantification, distillation, and others:

\noindent\textbf{Sparsification}: 
Gradients are sparse and only portions of gradients are actually useful for optimization~\cite{aji2017sparse}.
Consequently, sparsifiers including top-$k$ or random-$k$ are introduced in~\cite{aji2017sparse,lin2017deep,wangni2018gradient} to enable a compression rate of 1/100 and thereby greatly reducing communication costs.
Theoretical analyses in~\cite{sahu2021rethinking} also support these methods by formally showing that top-$k$ is the communication-optimal sparsifier under a limited communication budget. 
Nevertheless, no prior knowledge is used during sparsification, leading to high compressing errors~\cite{alistarh2018convergence} and affecting the convergence rate severely.

\noindent\textbf{Quantification}: 
It is well-known that full-precision (32-bit) data types are redundant in machine learning~\cite{denil2013predicting}.
Thus, transmitted data are cast into smaller data types before communicating~\cite{bernstein2018signsgd,alistarh2017qsgd,seide20141}, thereby reducing communication overhead.
However, since the 1-bit data type is the smallest, gradient quantification has limited flexibility and owns a compression rate of up to 1/32.

\noindent\textbf{Distillation}:
Wu~\cite{wu2022communication} used knowledge distillation to distill the local model to a smaller model before communicating. 
Nevertheless, it requires an extra public dataset for alignments between the teacher and the student, which not only incurs additional computational costs but also introduces domain gaps between the public dataset and the clients' local dataset.
Goetz~\cite{goetz2020federated} and Hu~\cite{hu2022fedsynth} propose to generate a synthetic dataset from the full dataset that has a smaller size and can be used to approximate gradients.
Still, the synthesis process requires simulating the multi-step optimization of model weights for multiple epochs, leading to not only high time and space complexity (\textit{i.e.}, gradients and intermediate model weights are calculated and stored many times), but also great instability and possible collapse (Figure~\ref{fig:pre-exp-one}) due to excessively deep back-propagation paths and the gradient explosion, especially for large models and datasets.

\noindent\textbf{Others}:
In~\cite{chen2019communication,zhou2021communication}, researchers alter the FL workflow to allow parallelism in the communication and the computation, without modifying the transmitted data.
Therefore, this parallelism can be used with other data-compressing algorithms combinedly to further accelerate the training process. Li~\cite{li2021communication} proposed utilizing compressed sensing in data compressing and decompressing.
However, the optimization of parameters in compressed sensing can be non-trivial. 

The proposed E-3SFC is quite different from the previous works mentioned above.
First, the proposed E-3SFC is more privacy-preserving than others, since the transmitted information is not even the model itself.
Under the same communication budget, E-3SFC empirically converges faster than the sparsification method (see Figure.~\ref{fig:accuracy-loss}).
Then, unlike distillation, the proposed E-3SFC imposes a lower computational burden since it does not employ public datasets and only utilizes single-step model optimization.
Additionally, E-3SFC possesses the merits of higher communication efficiency and flexibility.
The proposed E-3SFC can reduce the overall communication overhead during the uploading and downloading phases of FL with flexible communication budget schedulers.
In terms of flexibility, it can be seamlessly integrated into communication-computation parallelism and compressed sensing to to enhance its usability.

\begin{figure*}[!h]
    \centering
    \includegraphics[width=0.85\textwidth]{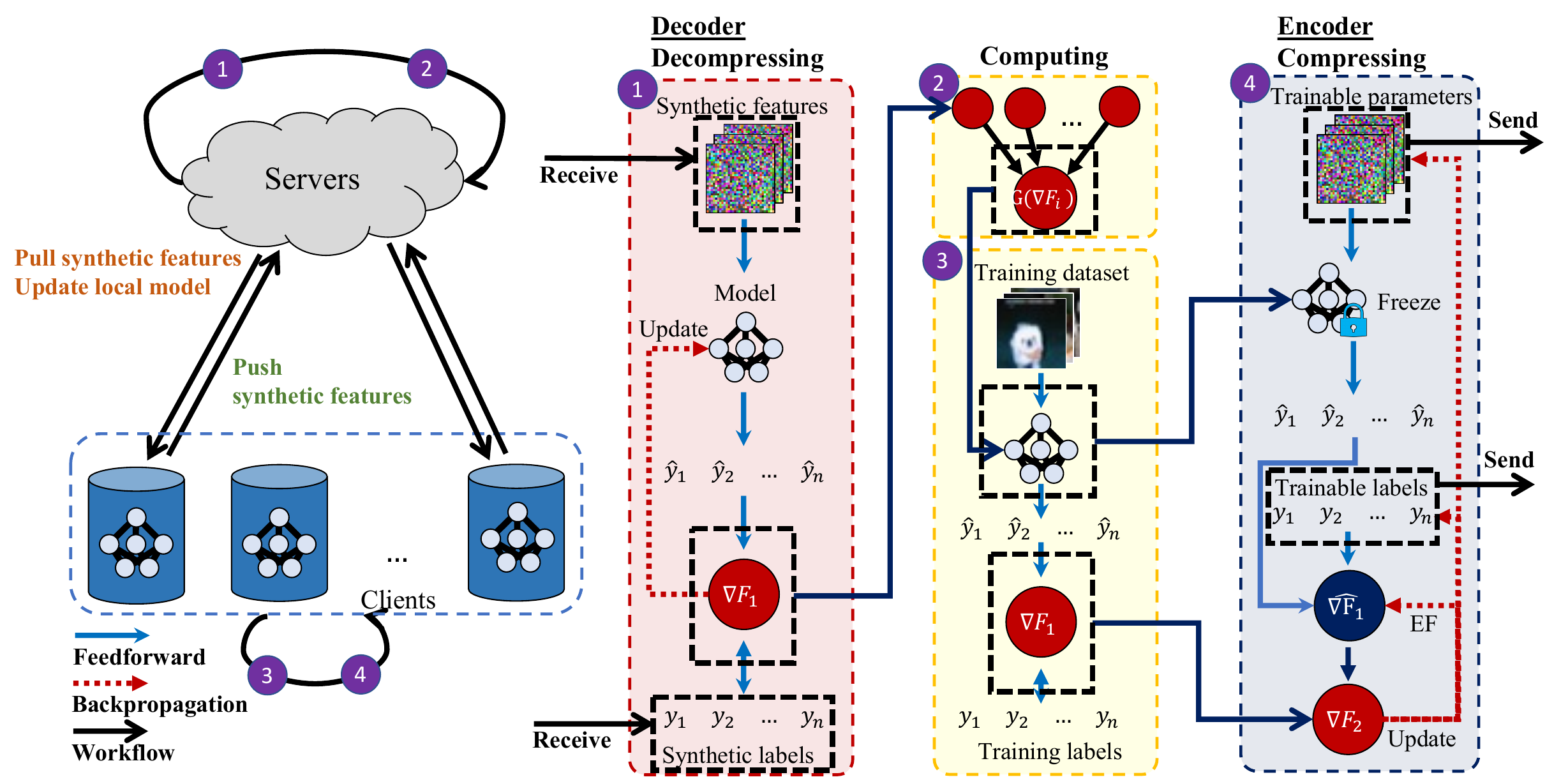}
    \caption{The general workflow of 3SFC. When compressing in \ding{185}, a set of trainable parameters and labels (i.e., synthetic features) will first be fed into the frozen local model to calculate model gradients. Then, calculated model gradients will be compared with real model gradients to optimize the synthetic features. When decompressing in \ding{182}, simply feed the local model with the received synthetic features and use the generated gradients to update the global model.}
    \label{fig:3sfc-arch}
\end{figure*}

\section{Problem Formulation}
\label{sec:problem-formulation}
Assume there are $N$ clients participating in a FL training process with a total of $T$ communication rounds, where the $i$-th client has a local dataset $D_i$ and a loss function $F(D_i, w_i)$ where $w_i$ is the weight of its model $M_i$. Note that in vanilla FL, all clients and servers share the same model architecture, \textit{i.e.}, $M_1 = M_2 = ... = M_N = M$. The objective of FL is to optimize Eq.~\ref{eq:fl-objective}:
\begin{equation}
    \min_{w \in \mathbb{R}^d} G(F(D_1, w), F(D_2, w), ..., F(D_N, w)),
    \label{eq:fl-objective}
\end{equation}
where $G(\cdot)$ is the linear aggregation function satisfying the sum of aggregation weights equals $1$. Typical aggregation functions include arithmetic mean and weighted average based on $p_i = |D_i|/\sum_j^N |D_j|$~\cite{mcmahan2017communication} where $|\cdot|$ is the size of the $\cdot$. The global model $w$ at the $t$-th communication round is updated by Eq.~\ref{eq:eq-3}:
\begin{align}
    w^{t+1} = G(w_1^{t}, w_2^{t}, ..., w_N^{t}) = w^{t} - G(\boldsymbol{g}_1^{t}, \boldsymbol{g}_2^{t}, ..., \boldsymbol{g}_N^{t}),
\label{eq:eq-3}
\end{align}
where $\boldsymbol{g}_i^t = w^t - w_i^t$ denotes the model weight differences after local training for $K$ rounds, and can be seen as accumulated gradients during local training. Generally, to reduce communication overhead, a compressor $\mathcal{C}$ is applied to each client's $\boldsymbol{g}_i^t$, so that the global model $w$ can be updated by Eq.~\ref{eq:eq-4}:
\begin{equation}
    w^{t+1} = w^{t} - G(\mathcal{C}(\boldsymbol{g}_1^t), \mathcal{C}(\boldsymbol{g}_2^t), ..., \mathcal{C}(\boldsymbol{g}_N^t)).
\label{eq:eq-4}
\end{equation}

As a result, the objective of communication compressing for client $i$ can be modeled as Eq.~\ref{eq:eq-5}:
\begin{equation}
    \mathcal{C}^* = \argmin || \mathcal{C}(\boldsymbol{g_i^t}) -\boldsymbol{g_i^t} ||^2 \text{ s.t. } ||\mathcal{C}(\boldsymbol{g_i^t})||_0 \leq B,
\label{eq:eq-5}
\end{equation}
where $B$ is the communication budget, constraining the maximum size of communication data at each communication round. 
Usually, the communication budget is strictly constrained for every communication round to support FL at scale.

Usually, an FL system contains servers located in data centers and a range of clients distributed across multiple areas.
The clients, such as mobile devices, are constrained by their data traffic plans.
On the other hand, many works~\cite{allen2019convergence, bejani2021systematic} have observed that the global model will converge progressively, and thus the gradient information from the early training phase should be prioritized over that from the later training phase.
To this end, we relax the communication budget by constraining only the total communication cost instead of the per-round communication cost.
Formally, the strict communication budget constraint can be relaxed as Eq.~\ref{eq:eq-relaxed-communication-constraint} shows.
\begin{equation}
\begin{aligned}
    &\mathcal{C}^* = \argmin \sum_{t}^{T} \sum_{i}^{N} || \mathcal{C}(\boldsymbol{g_i^t}) -\boldsymbol{g_i^t} ||^2 \\
    &\text{ s.t. } ||\mathcal{C}(\boldsymbol{g_i^t})||_0 \leq H(B, t, i), \sum_{t}^{T} \sum_{i}^{N} H(B, t, i) \leq TNB,
\end{aligned}
\label{eq:eq-relaxed-communication-constraint}
\end{equation}
where $H(\cdot)$ is a budget scheduler that returns a suitable communication budget for given $i$, $t$, and $B$.

Finally, letting $\boldsymbol{\epsilon}_i^t = ||\mathcal{C}(\boldsymbol{g_i^t}) - \boldsymbol{g_i^t}||$ denotes the compression error at time $t$, the error feedback can be utilized to optimize this error term by adding it to the $\boldsymbol{g_i^{t+1}}$. Thus, with error feedback, the global model can be updated as Eq.~\ref{eq:general-problem-optim}:
\begin{equation}
    \begin{aligned}
        w^{t+1} &= w^{t} - G\left(\mathcal{C}(\boldsymbol{g}_1^t + \boldsymbol{\epsilon}_1^{t}), ..., \mathcal{C}(\boldsymbol{g}_N^t + \boldsymbol{\epsilon}_N^{t})\right), \\
            \boldsymbol{\epsilon}_i^{t+1} &= \boldsymbol{g}_i^t + \boldsymbol{\epsilon}_i^{t} - \mathcal{C}(\boldsymbol{g}_{i}^t + \boldsymbol{\epsilon}_{i}^{t}).
    \end{aligned}
    \label{eq:general-problem-optim}
\end{equation}

\section{Methodology}
\label{sec:main-method}
In general, E-3SFC comprises three sub-components: a compressor 3SFC, a double-way compression algorithm, and a budget scheduler.
3SFC compresses clients' trained local models into compact features for uploading to the server to reduce communication overhead.
Moreover, the double-way compression algorithm allows 3SFC to compress the global model for distribution, reducing the communication costs of both model uploading and downloading.
Finally, the budget scheduler is introduced to dynamically determine $B$ for each communication round in 3SFC, enhancing system flexibility and efficiency.

\subsection{Compressing Clients' Local Models with 3SFC}
The general architecture of 3SFC is illustrated in Figure~\ref{fig:3sfc-arch}, where the training priors are injected in the compressing and decompressing process.
At each communication round, the $i$-th client first trains its local model using its local dataset. 
After training, accumulated gradients can be obtained by subtracting the global model weights from the latest local model weights. 
Then, according to the communication budget, the $i$-th client will utilize 3SFC to compress the accumulated gradients into synthetic features $D_{syn,i}^t$ using training priors, and sends $D_{syn,i}^t$ to the server. 
Formally, when compressing, 3SFC essentially converts $\boldsymbol{g}_i^t$ into synthetic features $D_{syn,i}^t$ and a scaling coefficient $s_i^t$ using client $i$'s training priors, \textit{i.e.}, $w^t$ and $F$, as both $w^t$ and $F$ are globally shared.
The objective of the compression can be described by Eq.~\ref{eq:encoder-objective-1}:
\begin{equation}
    \begin{aligned}
        \min_{D_{syn,i}^t, s_i^t} & \left(||s_i^t \nabla_{w^t} F(D_{syn,i}^t, w^t)  - \boldsymbol{g_i^t} - \boldsymbol{\epsilon}_i^t||^2 + \lambda {D_{syn,i}^t}^2\right) \\ 
    & \text{ s.t. } ||D_{syn,i}^t||_0 + 1 \leq B, 
    \end{aligned}
\label{eq:encoder-objective-1}
\end{equation}
where $\boldsymbol{g_i^t}$ denotes the differences between the global model $w^t$ and its latest local model, \textit{i.e.}, $\boldsymbol{g_i^t} = w^t - w_i^{t}$ for clients. $\lambda {D_{syn,i}^t}^2$ is an $\ell_2$ regularization term to constrain the sparsity of $D_{syn,i}^t$ for better stability. Note that here the global model $w^t$ is passed into $F(\cdot)$ instead of $w_i^t$, because $w^t$ is globally shared and is the initial weight of every client's local optimization process at each communication round. Since $\boldsymbol{g_i^t} + \boldsymbol{\epsilon_i^t}$ is fixed, $s_i^t$ can be derived from $\nabla_{w^t} F(D_{syn,i}^t, w^t)$ by ensuring that $\nabla_{w^t} F(D_{syn,i}^t, w^t)$ and $\boldsymbol{g_i^t} + \boldsymbol{\epsilon}_i^t$ are perpendicular to each other, as shown in Eq.~\ref{eq:s-compute}:
\begin{align}
\begin{split}
    s_i^t &= \frac{||\boldsymbol{g_i^t} + \boldsymbol{\epsilon}_i^t||}{||\nabla_{w^t} F(D_{syn,i}^t, w^t)||} cos(\theta) \\ &= \frac{(\boldsymbol{g_i^t} + \boldsymbol{\epsilon}_i^t) \cdot \nabla_{w^t} F(D_{syn,i}^t, w^t)}{||\nabla_{w^t} F(D_{syn,i}^t, w^t)||^2},
\label{eq:s-compute}
\end{split}
\end{align}
where $\theta$ is the angle between two $\boldsymbol{g_i^t} + \boldsymbol{\epsilon}_i^t$ and $\nabla_{w^t} F(D_{syn,i}^t, w^t)$. Consequently, the objective described in Eq.~\ref{eq:encoder-objective-1} is equivalent to the following optimization problem:
\begin{equation}
    \begin{aligned}
    \min_{D_{syn,i}^t} & \left(1 - |\frac{\nabla_{w^t} F(D_{syn,i}^t, w^t) \cdot (\boldsymbol{g_i^t} + \boldsymbol{\epsilon}_i^t)}{||\nabla_{w^t}  + \boldsymbol{\epsilon}_i^t||}| + \lambda {D_{syn,i}^t}^2\right) \\
    & \text{ s.t. } ||D_{syn,i}^t||_0 + 1 \leq B.
    \end{aligned}
\label{eq:dsyn-optim}
\end{equation}

Intuitively, the objective is to find synthetic features $D_{syn}^t$ that minimize the angle between generated gradients from $D_{syn}^t$ and $\boldsymbol{g_i^t}$.
This objective is similar to gradient leakage attacks~\cite{zhu2019deep, zhao2020idlg, yang2023gradient}, but since the clients are performing the attacks on themselves, it remains safe.
After solving Eq.~\ref{eq:dsyn-optim}, $s_i^t$ can be thus calculated by Eq.~\ref{eq:s-compute}.
To minimize the accumulated compression error with EF, $\boldsymbol{\epsilon}_i^t$ can be updated by Eq.~\ref{eq:general-problem-optim}. 
Finally, $D_{syn,i}^t$ and $s_i^t$ will be uploaded to the server to represent the local gradients of client $i$. 
When decompressing in the server, the gradients for the global model aggregation will be reconstructed using the received $D_{syn,i}^t$ and $s_i^t$ with the same training priors used for compressing, as shown in the following equation:
\begin{equation}
    \boldsymbol{g_i^t} + \boldsymbol{\epsilon}_i^t = s_i^t \nabla_{w^t} F(D_{syn,i}^t, w^t).
\end{equation}

\subsection{Double-way Compression}
The global model is distributed to all FL participants after weights are updated in the server, \textit{i.e.}, during the downloading phase. 
In vanilla FL, there are no training priors in the server for 3SFC to reduce the communication costs during model downloading and distribution.
Specifically, compressing $w^{t+1}$ in the server requires clients' local models $w_i^t$ as training priors, which are already compressed into $D_{syn,i}^t$ and $s_i^t$.
Moreover, since 3SFC is not a lossless compressor, it is infeasible for the server to recover $w_i^t$ based on $D_{syn,i}^t$ and $s_i^t$.

In order to further reduce communication costs, an algorithm for double-way compression is proposed.
First, $w^t$ is staged in the server after aggregating to serve as training priors for compressing $w^{t+1}$, as shown in Eq.~\ref{eq:dsyn-optim-server}:
\begin{equation}
    \begin{aligned}
    \min_{D_{syn}^{t+1}} & \left(1 - |\frac{\nabla_{w^t} F(D_{syn}^{t+1}, w^t) \cdot (w^t - w^{t+1} + \boldsymbol{\epsilon}^t)}{||\nabla_{w^t} F(D_{syn}^{t+1}, w^t)|| ||w^t - w^{t+1} + \boldsymbol{\epsilon}^t||}|\right. \\
    & \left. + \lambda {D_{syn}^{t+1}}^2\right)~\text{ s.t. } ||D_{syn}^{t+1}||_0 + 1 \leq B.
    \end{aligned}
\label{eq:dsyn-optim-server}
\end{equation}
By using $w^t$ as training priors for the server, we ensure that the compressed information is aligned with the clients’ perspectives, facilitating accurate decompression and reconstruction of $w^{t+1}$ on the client side.
Specifically, clients can easily recover the updated global model using the received $D_{syn}^{t}$ and $s^t$, as shown in the following equation:
\begin{equation}
    w^{t} = w^{t-1} - s^t \nabla_{w^{t-1}} F(D_{syn}^t, w^{t-1})
\end{equation}
Ensuring the same training priors are used is essential for the success of the decompression. 
Luckily, in FL, the global model $w^t$ is publicly available, and the objective functions for all participators are identical.
Consequently, the training priors are easily obtainable.

\begin{algorithm}[tb]
    \caption{E-3SFC (One Communication Round)}
    \label{alg:3sfc-algorithm}
    \textbf{Input}: local dataset $D_i$, learning rate $\eta^t$, accumulated gradient $\boldsymbol{\epsilon}_i^t$, regularization parameter $\lambda$\\
    \textbf{Parameter}: communication budget $B$, number of local iteration $K$, number of 3SFC iteration $S$, number of clients $N$, aggregation function $G$\\
    \textbf{Output}: global model $w^{t+1}$\\
    \textbf{Clients:}
    \begin{algorithmic}[1] %[1] enables line numbers
        \FOR{each client $i$ from $1$ to $N$ \textbf{in parallel}}
            \IF{\textit{double-way compression}}
                \STATE receive $D_{syn}^{t}$, $s^t$
                \STATE $w^{t} = w^{t-1} - s^t \nabla_{w^{t-1}} F(D_{syn}^t, w^{t-1})$
            \ELSE
                \STATE receive $w^{t}$
            \ENDIF
            \STATE $B^t = H(B, t, i)$
            \STATE initialize $D_{syn,i}^t$ where $||D_{syn,i}^t||_0 + 1 \leq B^t$
            \FOR{each local iteration $e$ from $1$ to $K$}
                \STATE $w_i^t = w_i^t - \eta^t \nabla_{w_i^t} F(D_i, w_i^t)$ \label{line:erm-optim}
            \ENDFOR
            \STATE $\boldsymbol{g}_i^t = w_i^t - w_i$
            \FOR{each $s$ from $1$ to $S$}
                \STATE Update $D_{syn,i}^{t}$ using Eq.~\ref{eq:dsyn-optim}\label{line:compression-optim}
            \ENDFOR
            \STATE $s_i^t = \frac{(\boldsymbol{g_i^t} + \boldsymbol{\epsilon}_i^t) \cdot \nabla_{w^t} F(D_{syn,i}^t, w^t)}{||\nabla_{w^t} F(D_{syn,i}^t, w^t)||^2}$
            \STATE $\boldsymbol{\epsilon}_i^{t+1} = \boldsymbol{\epsilon}_i^t + \boldsymbol{g}_i^t - s_i^t\nabla_{w^t} F(D_{syn,i}^t, w^t)$
            \RETURN $D_{syn,i}^t$, $s_i^t$, $\boldsymbol{\epsilon}_i^{t+1}$
        \ENDFOR
    \end{algorithmic}
    \textbf{Servers:}
    \begin{algorithmic}[1] %[1] enables line numbers
        \FOR{each client $i$ from $1$ to $N$}
            \STATE receive $D_{syn,i}^t$, $s_i^t$
            \STATE $\boldsymbol{g}_i^t = s_i^t \nabla_{w^t} F(D_{syn,i}^t, w^t)$
        \ENDFOR
        \STATE $w^{t+1} = w^{t} - G(\boldsymbol{g}_1^t, \boldsymbol{g}_2^t, ..., \boldsymbol{g}_N^t)$
        \IF{\textit{double-way compression}}
            \STATE initialize $D_{syn}^{t+1}$ where $||D_{syn}^{t+1}||_0 + 1 \leq B^{t}$
            \FOR{each $s$ from $1$ to $S$}
                \STATE Update $D_{syn}^{t+1}$ using Eq.~\ref{eq:dsyn-optim-server}
            \ENDFOR
            \STATE $s^t = \frac{(w^t - w^{t+1} + \boldsymbol{\epsilon}^{t+1}) \cdot \nabla_{w^t} F(D_{syn}^{t+1}, w^t)}{||\nabla_{w^t} F(D_{syn}^{t+1}, w^t)||^2}$
            \STATE $\boldsymbol{\epsilon}^{t+1} = \boldsymbol{\epsilon}^t + w^t - w^{t+1} - s^t \nabla_{w^t} F(D_{syn}^{t+1}, w^t)$
            \RETURN $D_{syn}^{t+1}$
        \ELSE
            \RETURN $w^{t+1}$
        \ENDIF
    \end{algorithmic}
\end{algorithm}

\subsection{Budget Scheduler}
The compressing process is not lossless and the compression efficiency decreases with the accumulation of the compressing error during the training process (As can be seen in latter experiments, \textit{i.e.}, Figure~\ref{fig:compression-efficiency}). 
This phenomenon motivates us to design a budget scheduler that assigns more communication budget in the early phase of the training under the relaxed constraint (\textit{i.e.}, Eq.~\ref{eq:eq-relaxed-communication-constraint}).
Formally, the budget scheduler $H(B, t, i)$ should satisfy the following equations to ensure that the overall communication overhead is unchanged:
\begin{equation}
    \underbrace{\frac{\sum_t^T H(B, t, i)}{T}}_{12a} = \underbrace{\frac{\sum_i^N H(B, t, i)}{N}}_{12b} = B; H(B, t, i) \geq 1. \nonumber
\end{equation}

Naturally, a $H(B, t, i)$ satisfying Eq.~12a automatically satisfies Eq.~12b by shifting its $t$ to $(t + iT/N) \mod T$. Thus, for simplicity, we will let $i = 0$ and discard Eq.~12b hereinafter. The objective of $H(\cdot)$ is to maximize the transferred information during the training process, as Eq.~\ref{eq:objective_budget_scheduler} illustrated.
\begin{equation}
    \begin{aligned}
        &\max_{H(\cdot)} \sum_t^T H(B, t, 0) \mathcal{E}(t) \\
        &\text{ s.t. } \sum_t^T H(B, t, 0) = BT \text{ and } H(B, t, 0) \geq 1,
    \end{aligned}
    \label{eq:objective_budget_scheduler}
\end{equation}
where $\mathcal{E}(t)$ is the compressing efficiency at time $t$. 
Although $\mathcal{E}(t)$ is unknown until the data is actually compressed, we empirically observe that $\mathcal{E}(t)$ is a decreasing function with respect to $t$ in Figure~\ref{fig:compression-efficiency}.
Thus, we let $H(B, t, 0)$ be parameterized by $w_H$, and altered the Eq.~\ref{eq:objective_budget_scheduler} to Eq.~\ref{eq:objective_budget_scheduler_sim} to efficiently find a closed-form solution of $H$ using constrained optimization methods.
\begin{equation}
\begin{aligned}
    &\max_{w_H} \sum_t^T  H(B, t, 0; w_H) (1 -\frac{t}{T}) - \tau \frac{|H(B, t, 0; w_H) - B|}{H(B, t, 0; w_H) - B}   \\ 
    &\text{ s.t. } \sum_t^T H(B, t, 0; w_H) = BT \text{ and } H(B, T, 0; w_H) \geq 1,
\end{aligned}
\label{eq:objective_budget_scheduler_sim}
\end{equation}

where $\tau$ is a hyper-parameter. Note that Eq.~\ref{eq:objective_budget_scheduler_sim} can be solved before the start of the training. Therefore, the employment of budget schedulers does not introduce any additional computational or communication overhead during the training process.

\subsection{Algorithm and Complexity Analysis}
The pseudocode of E-3SFC is presented in Algorithm~\ref{alg:3sfc-algorithm}.
During the training process, clients will solve two optimization problems instead of one compared to the vanilla FL method FedAvg~\cite{mcmahan2017communication}: the empirical risk minimization problem on the local dataset (Line~\ref{line:erm-optim}) and Eq.~\ref{eq:dsyn-optim} for compression (Line~\ref{line:compression-optim}). 
The solvers to these two problems are not nested, meaning the time complexity of E-3SFC and FedAvg is the same, \textit{i.e.}, $\mathcal{O}(NE(K + S))$.
In terms of the space complexity, E-3SFC additionally stores the $w^t$, $D_{syn,i}^t$, $s_i^t$ and $\boldsymbol{\epsilon}_i^t$, which are all fixed size parameters. Hence, E-3SFC shares the same space complexity, $\mathcal{O}(N)$, with FedAvg as well.

\section{Theoretical Analysis}
\label{sec:theoretical-analysis}
To unveil the theoretical properties of E-3SFC for enhanced explainability, this section provides the theoretical analysis of 3SFC under both strongly convex and non-convex settings, as E-3SFC includes 3SFC and inherits all of 3SFC’s theoretical properties.
To facilitate the analysis, the global epochs $T$ and the local epochs $K$ are concatenated and flattened. That is, there are in total $R=TK$ training epochs, and the communication happens only when $t \equiv 0 \mod{K}$. The detailed notations are listed in Table~\ref{tab:glossory}.

\begin{table}[ht]
    \centering
    \caption{The glossary of notations}
    \resizebox{0.45\textwidth}{!}{
    \begin{tabular}{cc}
        \hline
        Notation & Implication \\
        \hline
        $N$ & Number of total clients \\
        $M$ & Number of sampled clients for each epoch \\
        $S$ & Number of iterations to optimize $\mathcal{C}$ \\
        $T$ & Number of total global epochs \\
        $K$ & Number of local epochs \\
        $R$ & Number of flattened epochs, $R=TK$ \\
        $R_a$ & A set of flattened epochs for model aggregation \\
        $t_a$ & $t_a \in R_a$ \\
        $\cdot_{i}$ & $\cdot$ on $i^{th}$ client \\
        $\cdot^{t}$ & $\cdot$ at flattened epoch $t$ \\
        $D_{i}$ & Local dataset \\
        $F(\cdot)$ & Objective function \\
        $\tilde{F}(\cdot)$ & Approximated $F(\cdot)$ using $\xi_{i}$ sampled from $D_{i}$ \\
        $\mathcal{C}(\cdot)$ & Function for compressing \\
        $v_{i}^t$ & Local model, $v_i^0 = \boldsymbol{0}^n$ \\
        $w_{i}^t$ & Global model, $w_i^0 = \boldsymbol{0}^n$ \\
        $w^t$ & Aggregated global model, $w^t = \sum_{i=0}^N p_i w_i^t$ \\
        $\epsilon_i^t$ & $v_i^t - x_i^t$ \\
        $g_i^t$ & Compressed gradients, $g_i^t = \mathcal{C}(\eta^t \nabla F(w_i^t) + \epsilon_i^t, c_i^{t, S})$ \\
        $x_{i}^t$ & Accumulated full gradients, $x_i^t = \sum_{k=0}^t \eta^k \nabla F(w_i^k)$ \\
        $c_i^{t,s}$ & The parameters of $\mathcal{C}$ after optimizing $s$ iterations \\
        $\eta^t$ & Learning rate \\
        $p_i$ & The aggregation weight \\
        $\Delta^t$ & $||w^t - w^*||^2$ \\
        \hline
    \end{tabular}}
    \label{tab:glossory}
\end{table}

In order to present our theorem, some mild assumptions are made.

\begin{assumption}[Bounded Local Variance] The expectation of the difference between local gradients and global gradients is upper-bounded, \textit{i.e.},
    $\mathbb{E}||\nabla F(w)-\nabla F(w)||^{2}\leq \delta_{F}^2 + \lambda_{F}^{2}\mathbb{E}||\nabla F(w)||^{2}$.
\label{assump:1}
\end{assumption}

\begin{assumption}[Bounded Approximated Gradient] The expectation of the difference between local gradients and approximated local gradients using a subset of data is upper-bounded, \textit{i.e.}, 
    $\mathbb{E}||\nabla \tilde{F}(w)-\nabla F(w)||^2 \leq \sigma_F^2$.
\label{assump:2}
\end{assumption}

\begin{assumption}[Bounded Global Optimal Variance] The expectation of local gradients at the optimal solution is upper-bounded, \textit{i.e.}, 
    $\mathbb{E} ||\nabla F(w^*)||^2 \leq \sigma_*^2$.
\label{assump:3}
\end{assumption}

\begin{assumption}
    $F$ is expected-smooth, \textit{i.e.}, $\mathbb{E}_i ||\nabla F(w) - \nabla F(w^*)|| ^ 2 \leq 2 L (\mathbb{E}_i F(w) - \mathbb{E}_i F(w^*))$.
\label{assump:4}
\end{assumption}

\begin{assumption}[Unbiased Compression Optimization Variance Bound] The expectation of the difference between local gradients and compressed local gradients is upper-bounded, \textit{i.e.}, 
    $\mathbb{E} ||\eta \nabla F(w_i^t) - g_i^t||^2 \leq \eta^2 \kappa_S^2$,
where $\eta = \min\{\eta^{t_a}\}$.
\label{assump:5}
\end{assumption}

With these mild assumptions, We have Lemma~\ref{lemma:bounded-local-shift} and Lemma~\ref{lemma:bounded_local_approximation}. Details are in the Appendix.

\begin{lemma}[Bounded Local Shift]
    Under Assumption~\ref{assump:4}, with $\langle w^{t_a} - w^*, \nabla F(w^{t_a}) \rangle \geq F(w^{t_a}) - F(w^*) + \frac{\mu_F}{2} ||w^{t_a} - w^*|| ^ 2$, let $\tilde{\eta} = K \eta$, the upper-bound of local shift is:
    \begin{align}
        \mathbb{E}||w^{t_a + K} &- w^*||^2 \leq (1 - \frac{\tilde{\eta} \mu_F}{2}) \mathbb{E} ||w^{t_a} - w^*||^2 \nonumber \\
        &+ \frac{(\frac{2}{\mu_F \tilde{\eta}} + 3)K}{N} \sum_{i=0,k=0}^{N,K-1} \mathbb||g_i^{t_a + k} - \eta^k \nabla F(w^{t_a})||^2 \nonumber \\
        &+ (6L\tilde{\eta} - 2\tilde{\eta}) \mathbb{E}(F(w^{t_a}) - F(w^*)) \nonumber \\
        &+ 3\tilde{\eta}^2 \mathbb{E} ||\frac{1}{M} \sum_{i}^{M^t} \nabla F(w^{t_a}) - \nabla F(w^{t_a})||^2.
    \end{align}
\label{lemma:bounded-local-shift}
\end{lemma}

\begin{lemma}[Bounded Local Approximation] Under Assumption~\ref{assump:4} and Assumption~\ref{assump:5}, by choosing a proper $\tilde{\eta} \leq \frac{1}{6L}$, the upper-bound of local approximation is:
\begin{align}
    &~~~\mathbb{E}||g_i^{t_a+k} - \eta \nabla F(w^{t_a})||^2  \nonumber\\
    &\leq 3\eta^4 L e 2^{K+1} ((1 + 2K)||\nabla F(w^{t_a})||^2 + \dot{\delta}) + \eta^2 \dot{\delta},
\end{align}
where $\dot{\delta}:=3(\kappa_S^2+\sigma_F^2)$.
\label{lemma:bounded_local_approximation}
\end{lemma}

\begin{figure}
     \centering
     \begin{subfigure}[tb]{0.24\textwidth}
         \centering
         \includegraphics[width=\textwidth]{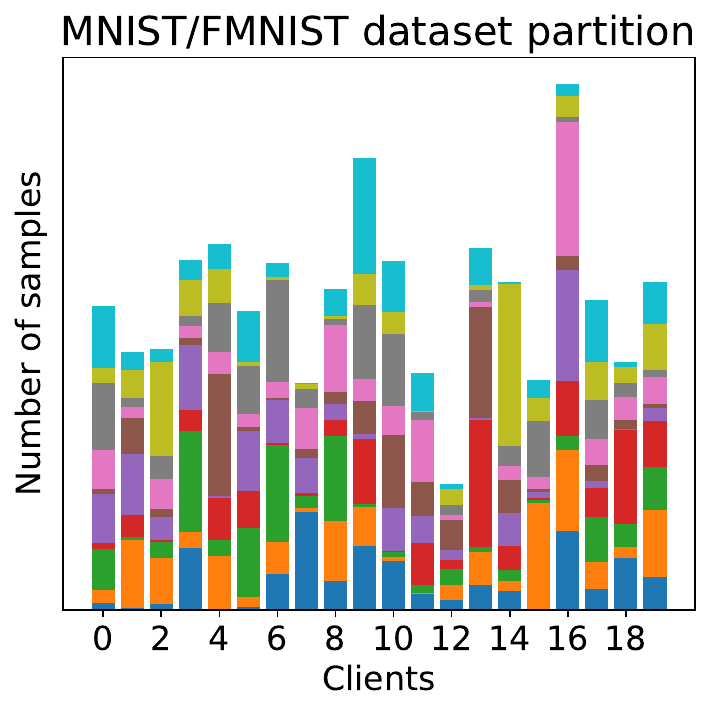}
     \end{subfigure}
     \hfill
     \begin{subfigure}[tb]{0.24\textwidth}
         \centering
         \includegraphics[width=\textwidth]{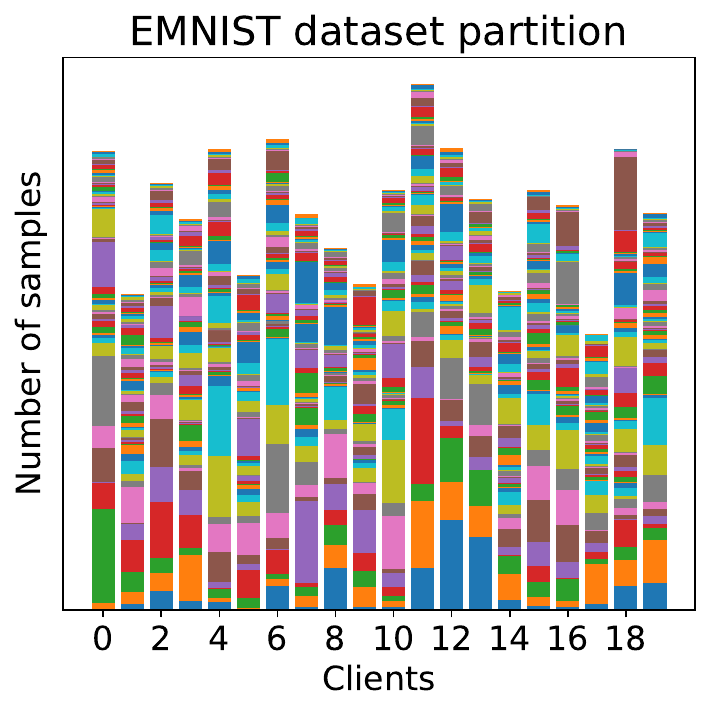}
     \end{subfigure}
     \\
     \begin{subfigure}[tb]{0.24\textwidth}
         \centering
         \includegraphics[width=\textwidth]{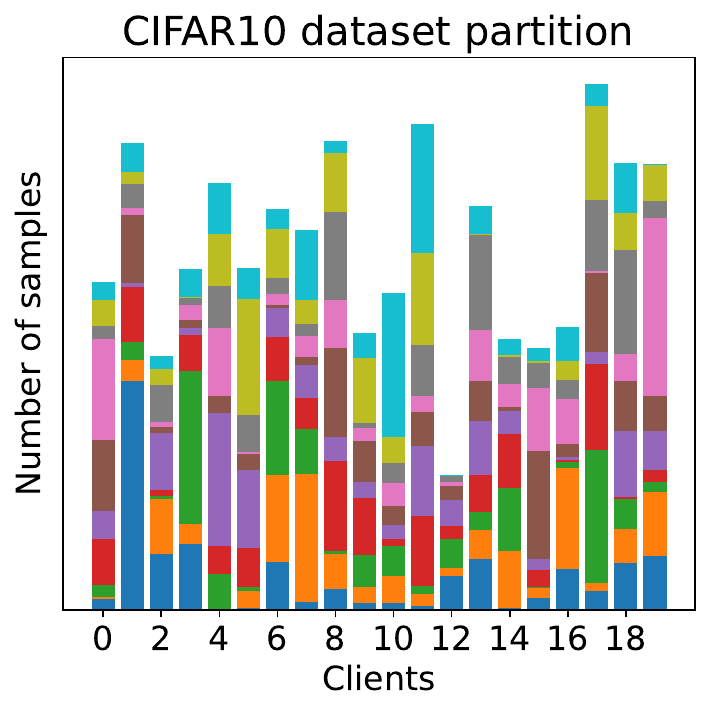}
     \end{subfigure}
     \hfill
     \begin{subfigure}[tb]{0.24\textwidth}
         \centering
         \includegraphics[width=\textwidth]{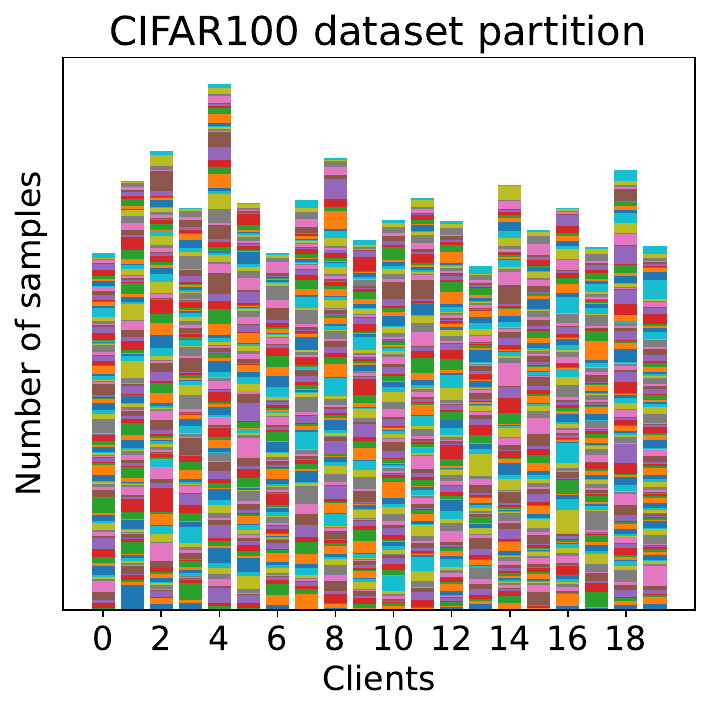}
     \end{subfigure}
        \caption{Illustration of our manual dataset partitions for 20 clients based on the Dirichlet distribution. Each bar represents a client, and different segments with different colors of a bar represent different labels. As can be seen, different clients have different dataset sizes and dataset distributions, and some clients only have some of the labels.}
        \label{fig:partition-illustration}
\end{figure}

With Lemma~\ref{lemma:bounded-local-shift} and Lemma~\ref{lemma:bounded_local_approximation}, the convergence rate of 3SFC under the strongly convex setting can be obtained.
\begin{theorem}
With Lemma~\ref{lemma:bounded-local-shift} and Lemma~\ref{lemma:bounded_local_approximation}, at flattened epoch $\tilde{\eta} R \geq \frac{2}{\mu_F}$, we have:
\begin{align}
    &~~~\mathcal{O}(F(\bar{w}) - F(w^*)) \nonumber \\
    &= \mathcal{O}(\mu_F e^{-\tilde{\eta} \mu_F \frac{R}{2} \Delta^0}) + \mathcal{O}(\frac{\kappa_S^2 + \sigma_F^2}{\mu_F}) \nonumber \\
    &~~~+ \mathcal{O}(\frac{N/M - 1}{N \mu_F R} \sigma_*^2) + \mathcal{O}(\frac{2^K L (\sigma_*^2 + \kappa_S^2 + \sigma_F^2)}{K^2 \mu_F^2 R^2})
\end{align}
\label{theorem:theorem_1}
\end{theorem}
\begin{corollary}
    Without aggregation noise, \textit{i.e.}, $N=M$, \textit{e.g.,} full client sampling, we have a quadratic convergence with total local epochs $R$.
\end{corollary}
\begin{remark}
    In the last term with $\sigma_*^2 + \kappa_S^2 + \sigma_F^2$, $S$ is chosen to maintain the same level variance of compression with the other two, in order for the compression not to affect the convergence rate too much. Due to the compression process with $S$, corresponding to $\kappa_{S}$, it shares the effect with approximate variance $\sigma_{F}^{2}$ on convergence radius and with optimal variance with $\sigma_{*}$ additionally on the quadratic term, which means $S$ is chosen to maintain the same level variance of compression with the other two.
\end{remark}
\begin{theorem}
\label{theo_2}
Under the non-convex setting and Lemma~\ref{lemma:bounded_local_approximation}, let Assumption~\ref{assump:1}, Assumption~\ref{assump:2} and Assumption~\ref{assump:3} hold, with $\tilde{\eta} \leq \frac{1}{L}$ and $\eta \leq \hat{\eta}_n := \frac{1}{(9LK + 27e2^{K+4})(1 + \lambda_F^2)}$, we have:
\begin{align}
     &~~~\mathcal{O}(\mathbb{E}||\nabla F(w^{t^*})||^2) \nonumber \\
     &= \mathcal{O}(\frac{\Delta^*}{\tilde{\eta}R}) + \mathcal{O}(\kappa_S^2+\sigma_F^2) + \mathcal{O}(\frac{{\Delta^*}^{\frac{1}{2}}L^{\frac{1}{2}}\delta_{F} (\frac{N/M-1}{N})^{\frac{1}{2}}}{K^{\frac{1}{2}}R^{\frac{1}{2}}}) \nonumber \\
     &~~~+ \mathcal{O}(\frac{{\Delta^*}^{\frac{2}{3}}2^{\frac{K}{3}}L^{\frac{2}{3}}}{KR^{\frac{2}{3}}} (\delta_{F}^{2}K + \kappa_S^2+\sigma_F^2)^{\frac{1}{3}}),
\end{align}
where $\Delta^* = \mathbb{E}(F(w^0) - F(w^*))$.
\label{theorem:theorem_2}
\end{theorem}
\begin{corollary}
\label{corollary:corollary_2}
    Tuning local epoch $K$, the main trade-off is between the first $\mathcal{O}(\frac{\Delta^{*}}{\tilde{\eta}R}) \propto \frac{1}{\eta K^{2} R}$ and last term $\propto \frac{2^{K/3}}{K^{2/3}}$.
\end{corollary}
\begin{remark}
    The compression process controlled by $S$ only affects the radius and part of the sub-linear term in $\kappa_S^2+\sigma_F^2$. With Corollary~\ref{corollary:corollary_2}, an optional strategy is to adapt $K$ decreasingly at the terminal stage of global training.
\end{remark}

\noindent\textbf{Discussion: }
The theorem~\ref{theorem:theorem_1} and~\ref{theorem:theorem_2} suggest that 3SFC has a $\mathcal{O}(\frac{1}{R})$/$\mathcal{O}(\frac{1}{R^2})$ speedup with/without aggregation noise under the strongly convex case, and a $\mathcal{O}(\frac{1}{R^{1/2}})$/$\mathcal{O}(\frac{1}{R^{2/3}})$ convergence rate with/without aggregation noise under the non-convex case.
The theoretical convergence results are similar to Top-$k$~\cite{alistarh2018convergence}, indicating that the proposed 3SFC and E-3SFC can achieve convergence.
The following empirical results demonstrate that 3SFC and E-3SFC obtain superior performance with much lower communication overhead, in practice.

\noindent\textbf{Insight: }
Striking a trade-off between convergence rate and communication overhead is challenging for many compression techniques.
For instance, the convergence rate of Top-$K$ is largely affected by the selection of $K$~\cite{alistarh2018convergence}.
Namely, increasing $K$ will also increase the model's convergence rate and the communication costs.
This holds true for quantification methods as well.
On the other hand, from our analysis, it is evident that the convergence rate of 3SFC is controlled by $\kappa_S$, which can be effectively reduced by increasing $S$.
In other words, 3SFC fixes communication overhead and balances convergence rate with computational overhead.
This unique property makes 3SFC an efficient compressor that achieves fast convergence while maintaining limited communication costs.

\begin{table}[tb]
  \centering
  \resizebox{0.8\linewidth}{!}{%
    \begin{tabular}{cccccc}
    \toprule
    \toprule
    \multirow{2}[2]{*}{Dataset+Model} & \multicolumn{1}{c}{\multirow{2}[2]{*}{FedAvg (1$\times$)}} & \multicolumn{2}{c}{FedSynth} & \multicolumn{1}{c}{\multirow{2}[2]{*}{3SFC (250$\times$)}} \\
      &       & \multicolumn{1}{c}{1$\times$} & \multicolumn{1}{c}{250$\times$} &        \\
    \midrule
    MNIST+MLP & 90.17 & 90.17   & 13.59 & 88.76 \\
    EMNIST+MLP & 61.08 & 61.08   & 1.92 & 54.94 \\
    FMNIST+MLP & 81.83 & 81.83   & 12.16 & 78.81 \\
    FMNIST+MnistNet & 85.73 & 85.73   & 13.18 & 81.79 \\
    \bottomrule
    \bottomrule
    \end{tabular}%
    }
  \caption{FedSynth is hard to converge, unlike the proposed 3SFC. As can be seen, the global model is barely optimized using FedSynth with a high compression ratio, while our proposed 3SFC achieves much higher performance. Consequently, FedSynth is not compared with 3SFC or E-3SFC in the latter experiments. These results validate our observations described in Section~\ref{sec:related-work}. These experiments were conducted with 10 clients after 200 epochs of training.}
  \label{tab:fedsynth-pre}%
\end{table}%

\section{Experiments Setup}
\label{sec:experiments}
In this section, we describe our experimental setups, including the selections of baselines, datasets, models, hyper-parameters, and more.

\noindent\textbf{Baselines:} We compare 3SFC and E-3SFC with five state-of-the-art methods: FedAvg~\cite{mcmahan2017communication}, DGC~\cite{lin2017deep}, signSGD with EF~\cite{bernstein2018signsgd}, z-sign~\cite{tang2024z} and STC~\cite{sattler2019robust}. Specifically, FedAvg is a traditional FL training method without any compression, DGC is considered as a \textit{state-of-the-art} in sparsification, signSGD, z-sign are quantification methods, and STC combines sparsification and quantification (\textit{i.e.}, STC sparsifies top-$k$ parameters and quantifies the others).

Note that previous work in the data distillation for FL realm (\textit{e.g.}, FedSynth~\cite{hu2022fedsynth} is considered as a \textit{state-of-the-art} in data distillation for FL realm to achieve communication efficient FL) is not compared in our experiments, as it hardly converges due to the instability and collapse described in Section~\ref{sec:related-work} with high compression ratio and large datasets and models, as Table~\ref{tab:fedsynth-pre} illustrated.

\noindent\textbf{Datasets:} Following the conventions of the community~\cite{sattler2019robust,zhou2021communication,bernstein2018signsgd}, six datasets including MNIST~\cite{deng2012mnist}, FMNIST~\cite{xiao2017fashion}, EMNIST~\cite{cohen2017emnist}, Cifar10, Cifar100~\cite{krizhevsky2009learning} red{and AgNews~\cite{zhang2015character}} are used in the experiments, where AgNews is an NLP dataset and the others are CV datasets. To simulate the Non-i.i.d. characteristic, all datasets are manually partitioned horizontally into multiple subsets based on the Dirichlet distribution~\cite{huang2005maximum}~($\alpha = 1.0$), which is commonly used in the FL setting~\cite{wang2020tackling,li2022federated}. Figure~\ref{fig:partition-illustration} illustrates our partitions. As can be seen, different clients own different datasets in terms of both quantity and category.
\begin{table*}[tb]
  \centering
  \resizebox{\linewidth}{!}{%
    \begin{tabular}{lrrrrrrrrrr}
    \toprule
    \toprule
    \multicolumn{1}{c}{\multirow{2}[4]{*}{Methods}} & \multicolumn{1}{c}{MNIST} & \multicolumn{1}{c}{EMNIST} & \multicolumn{2}{c}{FMNIST} & \multicolumn{3}{c}{Cifar10} & \multicolumn{2}{c}{Cifar100} & \multicolumn{1}{c}{AgNews} \\
\cmidrule{4-11}          & \multicolumn{1}{c}{MLP} & \multicolumn{1}{c}{MLP} & \multicolumn{1}{c}{MLP} & \multicolumn{1}{c}{MnistNet} & \multicolumn{1}{c}{ConvNet} & \multicolumn{1}{c}{ResNet} & \multicolumn{1}{c}{RegNet} & \multicolumn{1}{c}{ResNet} & \multicolumn{1}{c}{RegNet} & \multicolumn{1}{c}{TC} \\
    \midrule
    \midrule
    \multicolumn{11}{c}{10 Clients} \\
    \midrule
    FedAvg & 90.17$_{1.0\times}$ & 61.08$_{1.0\times}$ & 81.83$_{1.0\times}$ & 85.73$_{1.0\times}$ & 61.53$_{1.0\times}$ & 47.59$_{1.0\times}$ & 44.98$_{1.0\times}$ & 15.75$_{1.0\times}$ & 11.4$_{1.0\times}$ & 82.19$_{1.0\times}$ \\
    DGC   & 86.63$_{250.0\times}$ & 52.87$_{250.0\times}$ & 77.18$_{250.0\times}$ & 80.65$_{1333.3\times}$ & 61.51$_{10.4\times}$ & 21.13$_{3571.4\times}$ & 30.50$_{757.6\times}$ & 1.38$_{3571.4\times}$ & 3.22$_{757.6\times}$ & 80.22$_{1250.0\times}$ \\
    signSGD & 86.92$_{32.0\times}$ & 54.15$_{32.0\times}$ & 75.50$_{32.0\times}$ & 81.98$_{32.0\times}$ & 61.80$_{32.0\times}$ & 36.87$_{32.0\times}$ & 27.59$_{32.0\times}$ & 1.78$_{32.0\times}$ & 3.91$_{32.0\times}$ & 66.39$_{32.0\times}$ \\
    z-sign   & 86.69$_{32.0\times}$ & 52.87$_{32.0\times}$ & 75.49$_{32.0\times}$ & 82.17$_{32.0\times}$ & 61.81$_{\mathbf{32.0\times}}$ & 36.89$_{32.0\times}$ & 27.6$_{32.0\times}$ & 1.94$_{32.0\times}$ & 3.9$_{32.0\times}$ & 57.61$_{32.0\times}$ \\
    STC   & 88.48$_{32.0\times}$ & 52.58$_{32.0\times}$ & \textbf{80.16}$_{32.0\times}$ & \textbf{84.27}$_{32.0\times}$ & \underline{61.87}$_{\mathbf{32.0\times}}$ & \textbf{40.09}$_{32.0\times}$ & 35.68$_{32.0\times}$ & 0.88$_{32.0\times}$ & 4.16$_{32.0\times}$ & 56.18$_{32.0\times}$ \\
    \midrule
    3SFC  & \underline{88.76}$_{\mathbf{250.0\times}}$ & \underline{54.94}$_{\mathbf{250.0\times}}$ & 78.81$_{\mathbf{250.0\times}}$ & 81.79$_{\mathbf{1333.3\times}}$ & 61.82$_{10.4\times}$ & 25.67$_{\mathbf{3571.4\times}}$ & \underline{37.53}$_{\mathbf{757.6\times}}$ & \underline{4.66}$_{\mathbf{3571.4\times}}$ &  \underline{7.11}$_{\mathbf{757.6\times}}$ & \underline{81.64}$_{1250.0\times}$ \\
    E-3SFC  & \textbf{88.98}$_{\mathbf{250.0\times}}$ & \textbf{55.86}$_{\mathbf{250.0\times}}$ & \underline{79.06}$_{\mathbf{250.0\times}}$ & \underline{82.38}$_{\mathbf{1333.3\times}}$ & \textbf{62.69}$_{10.4\times}$ & \underline{25.88}$_{\mathbf{3571.4\times}}$ & \textbf{41.23}$_{\mathbf{757.6\times}}$ & \textbf{7.16}$_{\mathbf{3571.4\times}}$ &  \textbf{8.88}$_{\mathbf{757.6\times}}$ & \textbf{81.76}$_{1250.0\times}$ \\
    \midrule
    \midrule
    \multicolumn{11}{c}{20 Clients} \\
    \midrule
    FedAvg & 90.13$_{1.0\times}$ & 60.86$_{1.0\times}$ & 81.73$_{1.0\times}$ & 85.72$_{1.0\times}$ & 61.46$_{1.0\times}$ & 47.01$_{1.0\times}$ & 46.46$_{1.0\times}$ & 17.85$_{1.0\times}$ & 11.94$_{1.0\times}$ & 82.86$_{1.0\times}$ \\
    DGC   & 88.08$_{250.0\times}$ & 53.32$_{250.0\times}$ & 77.68$_{250.0\times}$ & 82.07$_{1333.3\times}$ & 61.15$_{10.4\times}$ & 25.42$_{3571.4\times}$ & 32.04$_{757.6\times}$ & 1.01$_{3571.4\times}$ & 5.01$_{757.6\times}$ & 81.3$_{1250.0\times}$ \\
    signSGD & 86.89$_{32.0\times}$ & 54.83$_{32.0\times}$ & 75.22$_{32.0\times}$ & 81.02$_{32.0\times}$ & 60.99$_{32.0\times}$ & \underline{36.73}$_{32.0\times}$ & 30.20$_{32.0\times}$ & \underline{8.24}$_{32.0\times}$ & 5.1$_{32.0\times}$ & 69.44$_{32.0\times}$ \\
    z-sign   & 86.69$_{2.0\times}$ & 54.95$_{32.0\times}$ & 75.2$_{32.0\times}$ & 81.14$_{32.0\times}$ & \underline{61.27}$_{\mathbf{32.0\times}}$ & 36.67$_{32.0\times}$ & 30.14$_{32.0\times}$ & 7.83$_{32.0\times}$ & 5.02$_{32.0\times}$ & 57.59$_{32.0\times}$ \\
    STC   & 88.89$_{32.0\times}$ & {55.12}$_{32.0\times}$ & {80.20}$_{32.0\times}$ & 81.98$_{32.0\times}$ & {61.25}$_{{32.0\times}}$ & \textbf{41.11}$_{32.0\times}$ & {37.48}$_{32.0\times}$ & {7.34}$_{32.0\times}$ & 4.99$_{32.0\times}$ & 57.01$_{32.0\times}$ \\
    \midrule
    3SFC  & \underline{89.18}$_{\mathbf{250.0\times}}$ & \underline{55.56}$_{\mathbf{250.0\times}}$ & \underline{80.13}$_{\mathbf{250.0\times}}$ & \underline{82.17}$_{\mathbf{1333.3\times}}$ & 60.44$_{10.4\times}$ & 30.49$_{\mathbf{3571.4\times}}$ & \underline{38.54}$_{\mathbf{757.6\times}}$ & 5.32$_{\mathbf{3571.4\times}}$ & \underline{7.64}$_{\mathbf{757.6\times}}$ & \underline{82.19}$_{1250.0\times}$ \\
    E-3SFC  & \textbf{89.31}$_{\mathbf{250.0\times}}$ & \textbf{56.59}$_{\mathbf{250.0\times}}$ & \textbf{80.39}$_{\mathbf{250.0\times}}$ & \textbf{82.47}$_{\mathbf{1333.3\times}}$ & \textbf{62.5}$_{10.4\times}$ & 30.67$_{\mathbf{3571.4\times}}$ & \textbf{42.42}$_{\mathbf{757.6\times}}$ & \textbf{8.73}$_{\mathbf{3571.4\times}}$ &  \textbf{8.95}$_{\mathbf{757.6\times}}$ & \textbf{82.25}$_{1250.0\times}$ \\
    \midrule
    \midrule
    \multicolumn{11}{c}{40 Clients} \\
    \midrule
    FedAvg & 90.03$_{1.0\times}$ & 61.38$_{1.0\times}$ & 81.62$_{1.0\times}$ & 85.59$_{1.0\times}$ & 60.36$_{1.0\times}$ & 46.53$_{1.0\times}$ & 45.97$_{1.0\times}$ & 16.79$_{1.0\times}$ & 10.47$_{1.0\times}$ & 82.73$_{1.0\times}$ \\
    DGC   & 87.75$_{250.0\times}$ & 54.25$_{250.0\times}$ & 76.45$_{250.0\times}$ & \underline{82.97}$_{1333.3\times}$ & 60.56$_{10.4\times}$ & 28.07$_{3571.4\times}$ & 33.79$_{757.6\times}$ & 0.94$_{3571.4\times}$ & 4.48$_{757.6\times}$ & 81.13$_{1250.0\times}$ \\
    signSGD & 86.98$_{32.0\times}$ & 55.83$_{32.0\times}$ & 75.46$_{32.0\times}$ & 81.24$_{32.0\times}$ & {61.02}$_{32.0\times}$ & \underline{37.79}$_{32.0\times}$ & 30.12$_{32.0\times}$ & {8.12}$_{32.0\times}$ & {4.75}$_{32.0\times}$ & 70.4$_{32.0\times}$\\
    z-sign   & 86.67$_{32.0\times}$ & 55.9$_{32.0\times}$ & 75.38$_{32.0\times}$ & 81.25$_{32.0\times}$ & 61.26$_{\mathbf{32.0\times}}$ & 37.62$_{32.0\times}$ & 30.0$_{32.0\times}$ & 7.62$_{32.0\times}$ & 4.83$_{32.0\times}$ & 56.2$_{32.0\times}$ \\
    STC   & {88.86}$_{32.0\times}$ & \underline{56.07}$_{32.0\times}$ & \underline{79.96}$_{32.0\times}$ & \textbf{83.10}$_{32.0\times}$ & 60.24$_{\mathbf{32.0\times}}$ & \textbf{41.28}$_{32.0\times}$ & {36.03}$_{32.0\times}$ & \underline{8.18}$_{32.0\times}$ & 4.14$_{32.0\times}$ & 58.06$_{32.0\times}$ \\
    \midrule
    3SFC  & \underline{88.86}$_{\mathbf{250.0\times}}$ & {55.95}$_{\mathbf{250.0\times}}$ & \underline{79.45}$_{\mathbf{250.0\times}}$ & 82.7$_{\mathbf{1333.3\times}}$ & \underline{61.45}$_{10.4\times}$ & 28.69$_{\mathbf{3571.4\times}}$ & \underline{38.35}$_{\mathbf{757.6\times}}$ & 5.60$_{\mathbf{3571.4\times}}$ & \underline{6.18}$_{\mathbf{757.6\times}}$ & \underline{82.03}$_{1250.0\times}$ \\
    E-3SFC  & \textbf{89.06}$_{\mathbf{250.0\times}}$ & \textbf{56.18}$_{\mathbf{250.0\times}}$ & \textbf{80.97}$_{\mathbf{250.0\times}}$ & 82.94$_{\mathbf{1333.3\times}}$ & \textbf{62.19}$_{10.4\times}$ & 28.67$_{\mathbf{3571.4\times}}$ & \textbf{43.62}$_{\mathbf{757.6\times}}$ & \textbf{8.55}$_{\mathbf{3571.4\times}}$ &  \textbf{7.23}$_{\mathbf{757.6\times}}$ & \textbf{82.11}$_{1250.0\times}$ \\
    \midrule
    \midrule
    \multicolumn{11}{c}{60 Clients} \\
    \midrule
    FedAvg  & 90.08$_{\mathbf{1.0\times}}$ & 58.54$_{\mathbf{1.0\times}}$ & 81.93$_{\mathbf{1.0\times}}$ & 85.49$_{\mathbf{1.0\times}}$ & 59.45$_{1.0\times}$ & 46.64$_{\mathbf{1.0\times}}$ & 45.45$_{\mathbf{1.0\times}}$ & 17.19$_{\mathbf{1.0\times}}$ &  11.51$_{\mathbf{1.0\times}}$ & 82.11$_{1.0\times}$ \\
    DGC  & 88.09$_{\mathbf{250.0\times}}$ & 47.05$_{\mathbf{250.0\times}}$ & 78.36$_{\mathbf{250.0\times}}$ & 82.07$_{\mathbf{1333.3\times}}$ & 59.49$_{10.4\times}$ & 26.61$_{\mathbf{3571.4\times}}$ & 33.97$_{\mathbf{757.6\times}}$ & 2.12$_{\mathbf{3571.4\times}}$ &  5.24$_{\mathbf{757.6\times}}$ & 81.08$_{1250.0\times}$ \\
    signSGD  & 87.12$_{\mathbf{32.0\times}}$ & \textbf{51.36}$_{\mathbf{32.0\times}}$ & 75.34$_{\mathbf{32.0\times}}$ & 80.84$_{\mathbf{32.0\times}}$ & 59.54$_{32.0\times}$ & \underline{38.24}$_{\mathbf{32.0\times}}$ & 30.95$_{\mathbf{32.0\times}}$ & 8.27$_{\mathbf{32.0\times}}$ &  5.36$_{\mathbf{32.0\times}}$ & 69.72$_{32.0\times}$ \\
    z-sign  & 86.6$_{\mathbf{32.0\times}}$ & 50.83$_{\mathbf{32.0\times}}$ & 75.38$_{\mathbf{32.0\times}}$ & 81.25$_{\mathbf{32.0\times}}$ & \textbf{61.31}$_{32.0\times}$ & 38.1$_{\mathbf{32.0\times}}$ & 30.89$_{\mathbf{32.0\times}}$ & 8.15$_{\mathbf{32.0\times}}$ &  5.31$_{\mathbf{32.0\times}}$ & 55.57$_{32.0\times}$ \\
    STC  & 89.01$_{\mathbf{32.0\times}}$ & 48.69$_{\mathbf{32.0\times}}$ & 80.23$_{\mathbf{32.0\times}}$ & \textbf{82.97}$_{\mathbf{32.0\times}}$ & 59.5$_{32.0\times}$ & \textbf{42.46}$_{\mathbf{32.0\times}}$ & 37.51$_{\mathbf{32.0\times}}$ & \textbf{9.08}$_{\mathbf{32.0\times}}$ &  5.29$_{\mathbf{32.0\times}}$ & 55.24$_{32.0\times}$ \\
    \midrule
    3SFC  & \underline{89.05}$_{\mathbf{250.0\times}}$ & 48.86$_{\mathbf{250.0\times}}$ & \underline{80.28}$_{\mathbf{250.0\times}}$ & 82.55$_{\mathbf{1333.3\times}}$ & 59.65$_{10.4\times}$ & 29.61$_{\mathbf{3571.4\times}}$ & \underline{39.84}$_{\mathbf{757.6\times}}$ & 7.02$_{\mathbf{3571.4\times}}$ &  \underline{6.93}$_{\mathbf{757.6\times}}$ & \underline{82.0}$_{1250.0\times}$ \\
    E-3SFC  & \textbf{89.06}$_{\mathbf{250.0\times}}$ & \underline{50.99}$_{\mathbf{250.0\times}}$ & \textbf{81.24}$_{\mathbf{250.0\times}}$ & \underline{82.67}$_{\mathbf{1333.3\times}}$ & \underline{60.45}$_{10.4\times}$ & 29.73$_{\mathbf{3571.4\times}}$ & \textbf{42.4}$_{\mathbf{757.6\times}}$ & \underline{8.74}$_{\mathbf{3571.4\times}}$ &  \textbf{8.28}$_{\mathbf{757.6\times}}$ & \textbf{82.56}$_{1250.0\times}$ \\
    \bottomrule
    \bottomrule
    \end{tabular}%
    }
    \caption{Comparison of test accuracy and compression ratio. We mark the best and second best performance by \textbf{bold} and \underline{underline}. Note that 3SFC, E-3SFC and DGC have much higher compression ratios compared to signSGD, z-sign and STC due to the limitation of quantification-based methods and the high compressing efficiency of 3SFC, E-3SFC and DGC. Consequently, while STC seems to perform well, 3SFC and E-3SFC achieves competing or better performance with a significantly lower communication budget. A dedicated comparison of 3SFC and STC is illustrated later to demonstrate the superiority of 3SFC in Section~\ref{sec:3sfc-stc-comp}.}
  \label{tab:accuracy-compare}%
\end{table*}%
\begin{figure}
     \centering
     \begin{subfigure}[tb]{0.24\textwidth}
         \centering
         \includegraphics[width=\textwidth]{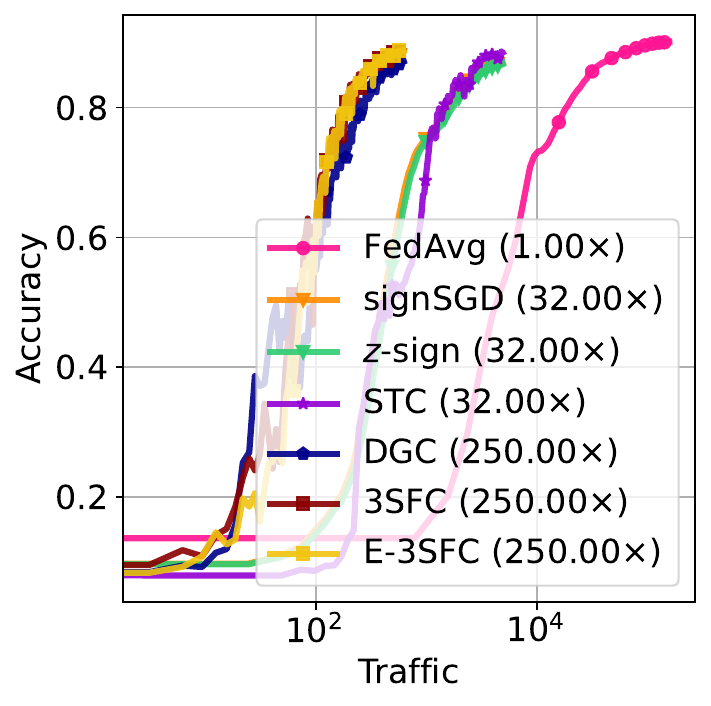}
         \caption{MLP trained on MNIST.}
     \end{subfigure}
     \hfill
     \begin{subfigure}[tb]{0.24\textwidth}
         \centering
         \includegraphics[width=\textwidth]{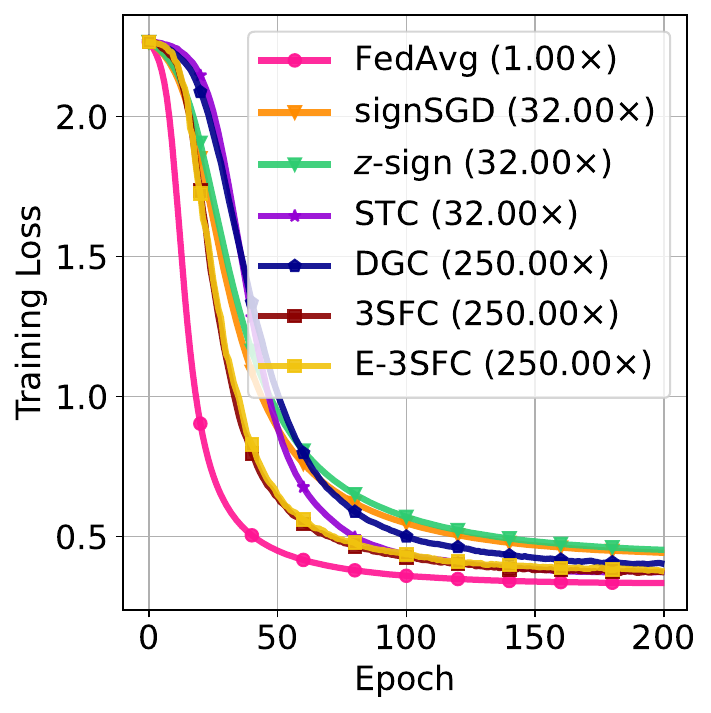}
         \caption{MLP trained on MNIST.}
     \end{subfigure}
     \\
     \begin{subfigure}[tb]{0.24\textwidth}
         \centering
         \includegraphics[width=\textwidth]{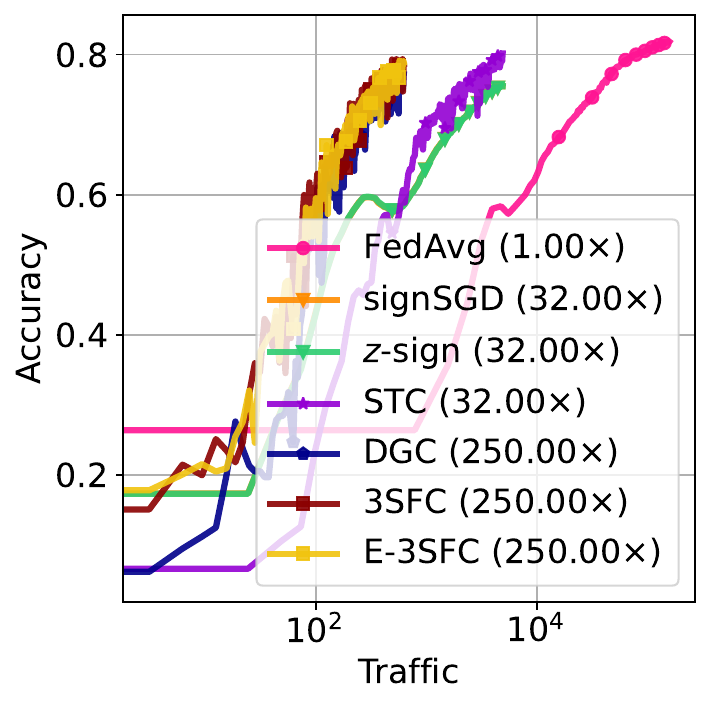}
         \caption{MLP trained on FMNIST.}
     \end{subfigure}
     \hfill
     \begin{subfigure}[tb]{0.24\textwidth}
         \centering
         \includegraphics[width=\textwidth]{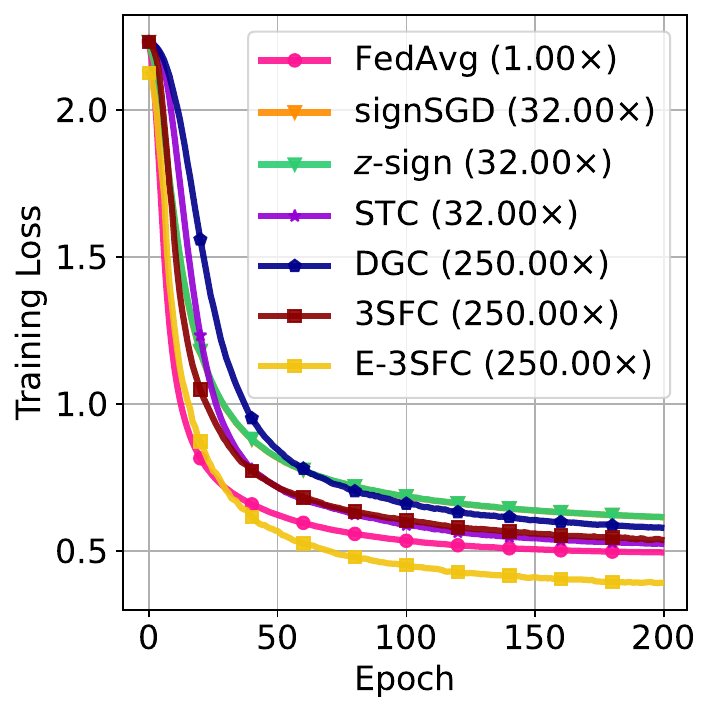}
         \caption{MLP trained on FMNIST.}
     \end{subfigure}
     \\
     \begin{subfigure}[tb]{0.24\textwidth}
         \centering
         \includegraphics[width=\textwidth]{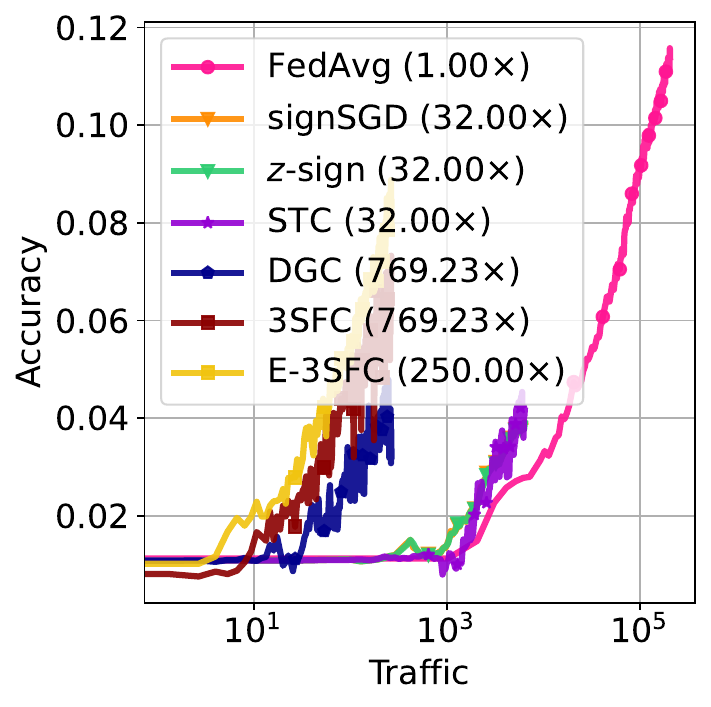}
         \caption{RegNet trained on Cifar100.}
     \end{subfigure}
     \hfill
     \begin{subfigure}[tb]{0.24\textwidth}
         \centering
         \includegraphics[width=\textwidth]{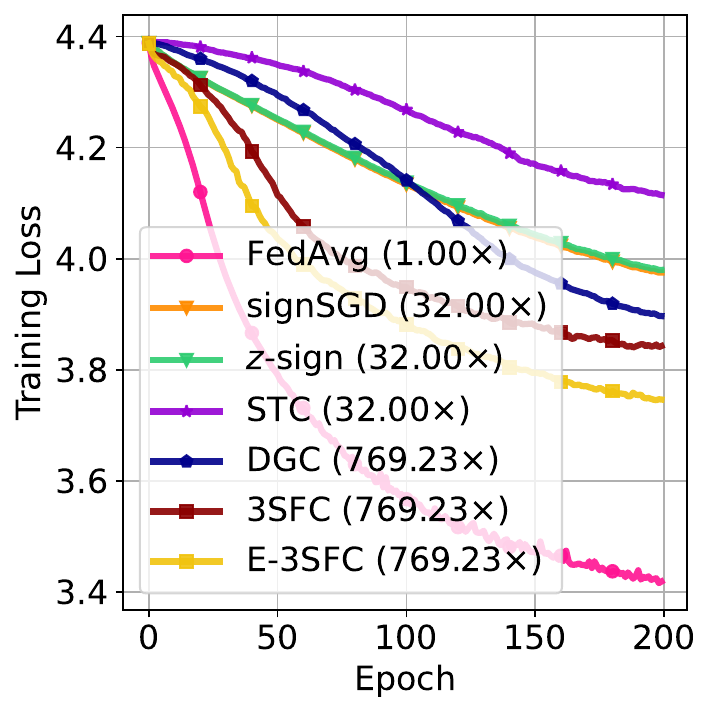}
         \caption{RegNet trained on Cifar100.}
     \end{subfigure}
        \caption{Test accuracy and training loss comparisons after 200 epochs of training. Compared to other methods, 3SFC and E-3SFC own the fastest convergence rate with respect to the amount of traffic communicated, with the highest compression ratio.}
        \label{fig:accuracy-loss}
\end{figure}
\begin{table}[tbp]
  \centering
  \resizebox{\linewidth}{!}{%
    \begin{tabular}{lrrr}
    \toprule
    \toprule
    Dataset+Model & \multicolumn{1}{l}{STC} & \multicolumn{1}{l}{3SFC ($2 \times B$)} & \multicolumn{1}{l}{3SFC ($4 \times B$)} \\
    \midrule
    \midrule
    \multicolumn{4}{c}{10 Clients} \\
    \midrule
    MNIST+MLP & 88.48$_{32.0\times}$ & \textbf{89.61}$_{\mathbf{125.0\times}}$ & \underline{89.58}$_{\underline{62.5\times}}$ \\
    EMNIST+MLP & 52.58$_{32.0\times}$ & \underline{58.20}$_{\mathbf{125.0\times}}$ & \textbf{59.55}$_{\underline{62.5\times}}$ \\
    FMNIST+MLP & 80.16$_{32.0\times}$ & \underline{80.31}$_{\mathbf{125.0\times}}$ & \textbf{80.63}$_{\underline{62.5\times}}$ \\
    FMNIST+MnistNet & \underline{84.27}$_{32.0\times}$ & 83.56$_{\mathbf{666.7\times}}$ & \textbf{84.30}$_{\underline{333.3\times}}$ \\
    Cifar10+ConvNet & 61.25$_{32.0\times}$ & \underline{61.65}$_{\underline{666.7\times}}$ & \textbf{61.87}$_{\mathbf{333.3\times}}$ \\
    Cifar10+Resnet & \textbf{40.09}$_{32.0\times}$ & 36.42$_{\mathbf{1785.7\times}}$ & \underline{39.54}$_{\underline{892.9\times}}$ \\
    Cifar10+Regnet & 35.68$_{32.0\times}$ & \underline{43.35}$_{\mathbf{378.8\times}}$ & \textbf{43.41}$_{\underline{189.4\times}}$ \\
    Cifar100+ResNet & 0.88$_{32.0\times}$ & \underline{8.81}$_{\mathbf{1785.7\times}}$ & \textbf{9.89}$_{\underline{892.9\times}}$ \\
    Cifar100+RegNet & 4.16$_{32.0\times}$ & \underline{9.46}$_{\mathbf{384.6\times}}$ & \textbf{9.52}$_{\underline{192.3\times}}$ \\
    AgNews+TC & 56.18$_{32.0\times}$ & \underline{81.49}$_{\mathbf{625.0\times}}$ & \textbf{81.58}$_{\underline{312.5\times}}$ \\
    \cellcolor{orange!20}Avg & \cellcolor{orange!20}50.44$_{32.0\times}$ & \cellcolor{orange!20}55.42$_{666.8\times}$ & \cellcolor{orange!20}56.04$_{333.4\times}$ \\
    \cellcolor{cyan!20}Std & \cellcolor{cyan!20}30.85 & \cellcolor{cyan!20}30.03 & \cellcolor{cyan!20}29.73 \\
    \midrule
    \midrule
    \multicolumn{4}{c}{20 Clients} \\
    \midrule
    MNIST+MLP & 88.89$_{32.0\times}$ & \underline{89.48}$_{\mathbf{125.0\times}}$ & \textbf{89.63}$_{\underline{62.5\times}}$ \\
    EMNIST+MLP & 55.12$_{32.0\times}$ & \underline{58.32}$_{\mathbf{125.0\times}}$ & \textbf{59.61}$_{\underline{62.5\times}}$ \\
    FMNIST+MLP & 80.20$_{32.0\times}$ & \underline{80.53}$_{\mathbf{125.0\times}}$ & \textbf{80.70}$_{\underline{62.5\times}}$ \\
    FMNIST+MnistNet & 81.98$_{32.0\times}$ & \underline{83.43}$_{\mathbf{666.7\times}}$ & \textbf{83.72}$_{\underline{333.3\times}}$ \\
    Cifar10+ConvNet & 61.87$_{32.0\times}$ & \textbf{63.08}$_{\mathbf{666.7\times}}$ & \underline{62.41}$_{\underline{333.3\times}}$ \\
    Cifar10+Resnet & \textbf{41.11}$_{32.0\times}$ & 34.50$_{\mathbf{1785.7\times}}$ & \underline{36.54}$_{\underline{892.9\times}}$ \\
    Cifar10+Regnet & 37.48$_{32.0\times}$ & \underline{43.76}$_{\mathbf{378.8\times}}$ & \textbf{45.08}$_{\underline{189.4\times}}$ \\
    Cifar100+ResNet & 7.34$_{32.0\times}$ & \underline{9.73}$_{\mathbf{1785.7\times}}$ & \textbf{11.18}$_{\underline{892.9\times}}$ \\
    Cifar100+RegNet & 4.99$_{32.0\times}$ & \underline{9.77}$_{\mathbf{384.6\times}}$ & \textbf{10.31}$_{\underline{192.3\times}}$ \\
    AgNews+TC & 57.01$_{32.0\times}$ & \underline{82.19}$_{\mathbf{625.0\times}}$ & \textbf{82.39}$_{\underline{312.5\times}}$ \\
    \cellcolor{orange!20}Avg & \cellcolor{orange!20}51.53$_{32.0\times}$ & \cellcolor{orange!20}55.33$_{666.8\times}$ & \cellcolor{orange!20}56.10$_{333.4\times}$ \\
    \cellcolor{cyan!20}Std & \cellcolor{cyan!20}29.26 & \cellcolor{cyan!20}29.97 & \cellcolor{cyan!20}29.52 \\
    \midrule
    \midrule
    \multicolumn{4}{c}{40 Clients} \\
    \midrule
    MNIST+MLP & 88.86$_{32.0\times}$ & \underline{89.32}$_{\mathbf{125.0\times}}$ & \textbf{89.49}$_{\underline{62.5\times}}$ \\
    EMNIST+MLP & 56.07$_{32.0\times}$ & \underline{58.76}$_{\mathbf{125.0\times}}$ & \textbf{59.95}$_{\underline{62.5\times}}$ \\
    FMNIST+MLP & 79.96$_{32.0\times}$ & \underline{80.27}$_{\mathbf{125.0\times}}$ & \textbf{80.73}$_{\underline{62.5\times}}$ \\
    FMNIST+MnistNet & 83.10$_{32.0\times}$ & \underline{83.74}$_{\mathbf{666.7\times}}$ & \textbf{84.12}$_{\underline{333.3\times}}$ \\
    Cifar10+ConvNet & 60.24$_{32.0\times}$ & \underline{61.32}$_{\underline{666.7\times}}$ & \textbf{61.55}$_{\mathbf{333.3\times}}$ \\
    Cifar10+Resnet & \textbf{41.28}$_{32.0\times}$ & \underline{37.47}$_{\mathbf{1785.7\times}}$ & 36.95$_{\underline{892.9\times}}$ \\
    Cifar10+Regnet & 36.03$_{32.0\times}$ & \underline{44.81}$_{\mathbf{378.8\times}}$ & \textbf{45.03}$_{\underline{189.4\times}}$ \\
    Cifar100+ResNet & 8.18$_{32.0\times}$ & \underline{10.41}$_{\mathbf{1785.7\times}}$ & \textbf{11.89}$_{\underline{892.9\times}}$ \\
    Cifar100+RegNet & 4.14$_{32.0\times}$ & \underline{7.99}$_{\mathbf{384.6\times}}$ & \textbf{8.89}$_{\underline{192.3\times}}$ \\
    AgNews+TC & 58.06$_{32.0\times}$ & \underline{82.06}$_{\mathbf{625.0\times}}$ & \textbf{82.18}$_{\underline{312.5\times}}$ \\
    \cellcolor{orange!20}Avg & \cellcolor{orange!20}51.59$_{32.0\times}$ & \cellcolor{orange!20}55.61$_{666.8\times}$ & \cellcolor{orange!20}56.03$_{333.4\times}$ \\
    \cellcolor{cyan!20}Std & \cellcolor{cyan!20}29.45 & \cellcolor{cyan!20}29.87 & \cellcolor{cyan!20}29.62 \\
    \midrule
    \midrule
    \multicolumn{4}{c}{60 Clients} \\
    \midrule
    MNIST+MLP & 89.01$_{32.0\times}$ & \underline{89.35}$_{\mathbf{125.0\times}}$ & \textbf{89.71}$_{\underline{62.5\times}}$ \\
    EMNIST+MLP & 48.69$_{32.0\times}$ & \underline{55.03}$_{\mathbf{125.0\times}}$ & \textbf{56.8}$_{\underline{62.5\times}}$ \\
    FMNIST+MLP & 80.23$_{32.0\times}$ & \underline{80.87}$_{\mathbf{125.0\times}}$ & \textbf{81.26}$_{\underline{62.5\times}}$ \\
    FMNIST+MnistNet & 82.97$_{32.0\times}$ & \underline{83.68}$_{\mathbf{666.7\times}}$ & \textbf{83.76}$_{\underline{333.3\times}}$ \\
    Cifar10+ConvNet & 59.5$_{32.0\times}$ & \underline{60.6}$_{\underline{666.7\times}}$ & \textbf{60.8}$_{\mathbf{333.3\times}}$ \\
    Cifar10+Resnet & \textbf{42.46}$_{32.0\times}$ & {37.53}$_{\mathbf{1785.7\times}}$ & \underline{37.62}$_{\underline{892.9\times}}$ \\
    Cifar10+Regnet & 37.51$_{32.0\times}$ & \underline{43.67}$_{\mathbf{378.8\times}}$ & \textbf{44.4}$_{\underline{189.4\times}}$ \\
    Cifar100+ResNet & 9.08$_{32.0\times}$ & \underline{9.54}$_{\mathbf{1785.7\times}}$ & \textbf{10.81}$_{\underline{892.9\times}}$ \\
    Cifar100+RegNet & 5.29$_{32.0\times}$ & \underline{9.57}$_{\mathbf{384.6\times}}$ & \textbf{10.17}$_{\underline{192.3\times}}$ \\
    AgNews+TC & 55.24$_{32.0\times}$ & \underline{82.16}$_{\mathbf{625.0\times}}$ & \textbf{82.21}$_{\underline{312.5\times}}$ \\
    \cellcolor{orange!20}Avg & \cellcolor{orange!20}50.99$_{32.0\times}$ & \cellcolor{orange!20}55.2$_{666.8\times}$ & \cellcolor{orange!20}55.72$_{333.4\times}$ \\
    \cellcolor{cyan!20}Std & \cellcolor{cyan!20}28.90 & \cellcolor{cyan!20}29.82 & \cellcolor{cyan!20}29.59 \\
    \bottomrule
    \bottomrule
    \end{tabular}%
    }
  \caption{Test accuracy and compression ratio comparisons of STC and 3SFC with different communication budgets. We mark the best and second best performance by \textbf{bold} and \underline{underline}. \colorbox{orange!20}{Avg} and \colorbox{cyan!20}{Std}: the average results and the standard deviation. 3SFC mostly achieves higher test accuracy while having a higher compression ratio, suggesting 3SFC compresses and decompresses the communication data more efficiently.}
  \label{tab:3sfc-stc-comp}%
\end{table}%

\noindent\textbf{Models:} To cover both simple and complicated learning problems, six models including Multi-Layer Perceptron (MLP), MnistNet, ConvNet, ResNet~\cite{he2016deep}, RegNet~\cite{radosavovic2020designing} and TC are used in the experiments. Here, MnistNet has two convolutional layers and two linear layers, ConvNet has four convolutional layers and one linear layer, and TC is a RNN with text embeddings specialized for NLP. Additionally, for ResNet and RegNet, all batch normalization layers~\cite{ioffe2015batch} and dropout layers~\cite{srivastava2014dropout} are deleted from the model as their parameters are not trainable, and their values cannot be aggregated simply by weighted averaging~\cite{wang2023batch}. This simplification has also been used in previous studies\cite{sattler2019robust,zhou2021communication}.

\noindent\textbf{Implementaion Details:} All experiments are evaluated on a simulated 60 clients cluster under a Non-IID environment in order to simulate the real-world scenario. The CUDA version is 11.4, the Python version is 3.9.15 and the PyTorch version is 1.13.0. The code of the experiments was developed with Pytorch and Pytorch Distributed RPC\footnote{https://pytorch.org/docs/stable/rpc.html} Framework with GLOO backend\footnote{https://github.com/facebookincubator/gloo} from scratch. NCCL backend was not used in our experiments due to its limited support for many edge devices without Nvidia GPU. The learning rate is set to 0.01, the batch size $B$ is set to $256$, local iteration $K$ is set to $5$, 3SFC iteration $S$ is set to $10$, and $\lambda$ is set to 0 for no regularization. For better readability, the compression ratio is used to measure the communication efficiency, which is defined as Eq.~\ref{eq:define-compression-rate}.
\begin{equation}
    \text{Comp. Rate} = \frac{\text{Comp. Size}}{\text{Uncomp. Size}} = \frac{1}{\text{Comp. Ratio}}.
    \label{eq:define-compression-rate}
\end{equation}

\section{Experimental Analysis}
\label{sec:experimental-analysis}
In this section, we present and analyze our experimental results. 
We begin by comparing the generalization performance and compression ratio of various compressors across diverse settings. 
Then, we report the test performance of different budget schedulers to show the effectiveness of our proposed schedulers.
Our findings indicate that both 3SFC and E-3SFC generally outperform STC, though STC achieves the best performance in a few specific instances. 
To further illustrate the superiority of 3SFC, we provide a detailed comparison with STC, implicitly highlighting the advantages of E-3SFC. 
We also visualize the compression efficiency of the different compressors.
Finally, we conduct ablation studies to demonstrate the robustness of E-3SFC.
\subsection{Overall Performance Comparisons}
We first compare the final test accuracy of 3SFC and E-3SFC with other competing methods after 200 epochs of training. In terms of the compression ratio, we set DGC to be the same as 3SFC and E-3SFC for all experiments for fair comparisons, because DGC is a sparsification-based method that has the flexibility to set an extremely high compression ratio. In contrast, for quantification-based methods like signSGD, z-sign and STC, we leave their compression ratio to be the maximal, \textit{i.e.}, $32\times$, and will later do dedicated evaluations between 3SFC and them in Section~\ref{sec:3sfc-stc-comp}.

The comprehensive test accuracy comparison results of 3SFC, E-3SFC and other methods are shown in Table~\ref{tab:accuracy-compare}, where the corresponding compression ratio is annotated as subscripts. It can be observed that under the same compression ratio, 3SFC yields higher test accuracy consistently compared to DGC after training, suggesting that 3SFC brings a faster convergence rate to the model training when the communication budget is equally limited. On the other hand, 3SFC still achieves comparable model performance compared to signSGD, z-sign and STC while communicates much less frequently, (\textit{i.e.}, 100$\times$ less for ResNet). Figure~\ref{fig:accuracy-loss} further validates the effectiveness of 3SFC by visualizing curves of test accuracy and training loss.
Additionally, E-3SFC outperforms 3SFC under the same compression ratio across all settings.
This is because, to maintain the same compression ratio, $B$ is slightly increased due to the communication overhead reduction in the model downloading phase, indicating that E-3SFC achieves higher compression efficiency compared to 3SFC.

\begin{table*}[tb]
  \centering
  \resizebox{0.93\linewidth}{!}{%
    \begin{tabular}{lllrrrrrrrrrrrr}
    \toprule
    \toprule
\multicolumn{3}{c}{Components} & \multicolumn{1}{c}{MNIST} & \multicolumn{1}{c}{EMNIST} & \multicolumn{2}{c}{FMNIST} & \multicolumn{3}{c}{Cifar10} & \multicolumn{2}{c}{Cifar100} & \multicolumn{1}{c}{AgNews} & \multicolumn{2}{c}{Statistics} \\
\cmidrule{1-15} \multicolumn{1}{c}{BS} & \multicolumn{1}{c}{DWC} & \multicolumn{1}{c}{EF} & \multicolumn{1}{c}{MLP} & \multicolumn{1}{c}{MLP} & \multicolumn{1}{c}{MLP} & \multicolumn{1}{c}{MnistNet} & \multicolumn{1}{c}{ConvNet} & \multicolumn{1}{c}{ResNet} & \multicolumn{1}{c}{RegNet} & \multicolumn{1}{c}{ResNet} & \multicolumn{1}{c}{RegNet} & \multicolumn{1}{c}{TC} & \multicolumn{1}{c}{\cellcolor{orange!20}Avg} & \multicolumn{1}{c}{\cellcolor{cyan!20}Std} \\
    \midrule
    \midrule
    \multicolumn{13}{c}{10 Clients} \\
    \midrule
    & & & 45.80 & 23.97 & 57.46 & 73.24 & 44.95 & 23.13 & 25.59 & 1.70& 2.35& 66.54 & \cellcolor{orange!20}37.47 & \cellcolor{cyan!20}25.72 \\
    \checkmark & & & 60.05 & 20.90 & 70.92 & 77.01 & 52.19 & 25.32 & 31.97 & 2.03  & 3.74 & 71.31 & \cellcolor{orange!20}41.54 & \cellcolor{cyan!20}28.35 \\
    & \checkmark & & 60.16 & 21.24 & 70.87 & 77.39 & 53.21 & 26.17 & 33.79 & 4.08  & 4.15 & 71.32 & \cellcolor{orange!20}42.23 & \cellcolor{cyan!20}27.93 \\
    & & \checkmark & 88.76 & 54.94 & 78.81 & 81.79 & 61.82 & 25.67 & 37.53 & 4.66& 7.11& 81.64 & \cellcolor{orange!20}52.27 & \cellcolor{cyan!20}31.79 \\
    & \checkmark & \checkmark & 88.98 & 55.86 & 79.06 & 82.38 & 62.69 & 25.88 & 41.23 & 7.16 &  8.88 & 81.76 & \cellcolor{orange!20}53.39 & \cellcolor{cyan!20}31.06 \\
    \checkmark & & \checkmark & 88.91 & 55.14 & 79.15 & 81.98 & 61.81 & 25.88 & 38.58 & 4.73 &  7.29 & 81.72 & \cellcolor{orange!20}52.52 & \cellcolor{cyan!20}31.76 \\
    \midrule
    \midrule
    \multicolumn{13}{c}{20 Clients} \\
    \midrule
    & & & 67.07 & 19.70 & 60.08 & 74.50 & 46.25 & 23.73 & 30.62 & 2.31& 2.95& 66.1 & \cellcolor{orange!20}39.33 & \cellcolor{cyan!20}27.07 \\
    \checkmark & & & 72.08 & 23.29 & 71.73 & 77.01 & 50.72 & 24.42 & 35.06 & 2.32 & 4.22 & 71.46 & \cellcolor{orange!20}43.23 & \cellcolor{cyan!20}29.16 \\
    & \checkmark & & 71.85 & 23.89 & 70.98 & 77.23 & 51.92 & 25.01 & 38.08 & 4.24 & 4.79 & 71.76 & \cellcolor{orange!20}43.97 & \cellcolor{cyan!20}28.59 \\
    & & \checkmark & 89.18 & 55.56 & 80.13 & 82.17 & 60.44 & 30.49 & 38.54 & 5.32& 7.64& 82.19 & \cellcolor{orange!20}53.16 & \cellcolor{cyan!20}31.36 \\
    & \checkmark & \checkmark & 89.31 & 56.59 & 80.39 & 82.47 & 62.5 & 30.67 & 42.42 & 8.73 &  8.95 & 82.25 & \cellcolor{orange!20}54.43 & \cellcolor{cyan!20}30.52 \\
    \checkmark & & \checkmark & 89.28 & 55.78 & 80.87 & 82.24 & 60.57 & 30.7 & 39.14 & 5.52 &  7.72 & 82.11 & \cellcolor{orange!20}53.39 & \cellcolor{cyan!20}31.36 \\
    \midrule
    \midrule
    \multicolumn{13}{c}{40 Clients} \\
    \midrule
    & & & 48.30 & 25.12 & 63.49 & 75.75 & 47.42 & 23.00  & 29.17 & 1.33 & 2.77 & 61.87 & \cellcolor{orange!20}39.82 & \cellcolor{cyan!20}27.22 \\
    \checkmark & & & 66.39 & 23.66 & 70.25 & 77.72 & 50.66 & 25.11 & 34.91 & 2.07 & 3.45 & 67.68 & \cellcolor{orange!20}42.19 & \cellcolor{cyan!20}28.19 \\
    & \checkmark & & 66.86 & 24.17 & 69.89 & 77.94 & 51.51 & 25.01 & 38.32 & 5.25 & 4.13 & 67.64 & \cellcolor{orange!20}43.07 & \cellcolor{cyan!20}23.55 \\
    & & \checkmark & 88.86 & 55.95 & 79.45 & 82.70 & 61.45 & 28.69 & 38.35 & 5.60& 6.18& 82.03 & \cellcolor{orange!20}52.92 & \cellcolor{cyan!20}31.68 \\
    & \checkmark & \checkmark & 89.06 & 56.18 & 80.97 & 82.94 & 62.19 & 28.67 & 43.62 & 8.55 &  7.23 & 82.11 & \cellcolor{orange!20}54.15 & \cellcolor{cyan!20}31.01 \\
    \checkmark & & \checkmark & 88.95 & 56.01 & 79.49 & 82.83 & 61.59 & 28.88 & 38.56 & 5.77 &  6.24 & 82.19 & \cellcolor{orange!20}53.05 & \cellcolor{cyan!20}31.66 \\
    \midrule
    \midrule
    \multicolumn{13}{c}{60 Clents} \\
    \midrule
    & & & 70.92 & 4.48 & 64.33 & 75.75 & 45.31 & 21.67  & 30.34 & 1.01 & 2.46 & 61.94 & \cellcolor{orange!20}37.82 & \cellcolor{cyan!20}29.64 \\
    \checkmark & & & 75.19 & 8.44 & 71.03 & 76.97 & 48.95 & 24.29 & 34.55 & 1.40 & 3.22 & 66.94 & \cellcolor{orange!20}41.09 & \cellcolor{cyan!20}30.69 \\
    & \checkmark & & 74.96 & 7.84 & 72.59 &  77.18 & 49.05 & 24.62 & 36.17 & 2.09 & 4.92 & 66.88 & \cellcolor{orange!20}41.63 & \cellcolor{cyan!20}30.54 \\
    & & \checkmark & 89.05 & 48.86 & 80.28 & 82.55 & 59.65 & 29.61 & 39.84 & 7.02 & 6.93 & 82.0 & \cellcolor{orange!20}52.57 & \cellcolor{cyan!20}31.22 \\
    & \checkmark & \checkmark & 89.06 & 50.99 & 81.24 & 82.67 & 60.45 & 29.73 & 42.4 & 8.74 &  8.28 & 82.56 & \cellcolor{orange!20}53.61 & \cellcolor{cyan!20}30.76 \\
    \checkmark & & \checkmark & 89.17 & 48.93 & 80.31 & 82.59 & 59.74 & 29.77 & 39.91 & 7.12 &  6.99 & 82.18 & \cellcolor{orange!20}52.67 & \cellcolor{cyan!20}31.22 \\
    \bottomrule
    \bottomrule
    \end{tabular}%
    }
  \caption{The ablation study of E-3SFC equipped with different sub-components, \textit{i.e.}, Budget Scheduler (BS), Double-way Compression (DWC) and EF. The results validate the effectiveness of each component in E-3SFC. \colorbox{orange!20}{Avg} and \colorbox{cyan!20}{Std}: the average results and the standard deviation.}
  \label{tab:ablation-e3sfc}%
\end{table*}%

\begin{table*}[tb]
  \centering
  \resizebox{0.95\linewidth}{!}{%
    \begin{tabular}{lrrrrrrrrrrrr}
    \toprule
    \toprule
    \multicolumn{1}{c}{\multirow{2}[4]{*}{Methods}} & \multicolumn{1}{c}{MNIST} & \multicolumn{1}{c}{EMNIST} & \multicolumn{2}{c}{FMNIST} & \multicolumn{3}{c}{Cifar10} & \multicolumn{2}{c}{Cifar100} & \multicolumn{1}{c}{AgNews} & \multicolumn{2}{c}{Statistics} \\
\cmidrule{4-13}          & \multicolumn{1}{c}{MLP} & \multicolumn{1}{c}{MLP} & \multicolumn{1}{c}{MLP} & \multicolumn{1}{c}{MnistNet} & \multicolumn{1}{c}{ConvNet} & \multicolumn{1}{c}{ResNet} & \multicolumn{1}{c}{RegNet} & \multicolumn{1}{c}{ResNet} & \multicolumn{1}{c}{RegNet} & \multicolumn{1}{c}{TC} & \multicolumn{1}{c}{\cellcolor{orange!20}Avg} & \multicolumn{1}{c}{\cellcolor{cyan!20}Std} \\
    \midrule
    \midrule
    \multicolumn{13}{c}{10 Clients} \\
    \midrule
    3SFC w/ EF  & 88.76 & 54.94 & 78.81 & 81.79 & 61.82 & 25.67 & 37.53 & 4.66 & 7.11 & 81.64 & \cellcolor{orange!20}52.27 & \cellcolor{cyan!20}31.79 \\
    3SFC w/o EF & 45.80 & 23.97 & 57.46 & 73.24 & 44.95 & 23.13 & 25.59 & 1.70 & 2.35 & 66.54 & \cellcolor{orange!20}37.47 & \cellcolor{cyan!20}25.72 \\
    3SFC w/ EF ($2\times B$)   & 89.61 & 58.20 & 80.31 & 83.56 & 63.08 & 36.42 & 43.35 & 8.81 & 9.46 & 81.49 & \cellcolor{orange!20}55.42 & \cellcolor{cyan!20}30.03 \\
    3SFC w/ EF ($4\times B$)   & 89.58 & 59.55 & 80.63 & 84.30 & 62.41 & 39.54 & 43.41 & 9.89 & 9.52 & 81.58 & \cellcolor{orange!20}56.04 & \cellcolor{cyan!20}29.73 \\
    3SFC w/ EF ($K=1$) & 69.39 & 31.52 & 65.00  & 78.07 & 52.07 & 20.01 & 28.71 & 1.04 & 3.66 & 57.52 & \cellcolor{orange!20}40.69 & \cellcolor{cyan!20}27.53 \\
    3SFC w/ EF ($K=10$) & 89.61 & 60.75 & 82.12 & 83.83 & 63.33 & 30.99 & 39.08 & 4.94 & 7.78 & 85.41 & \cellcolor{orange!20}54.78 & \cellcolor{cyan!20}32.24 \\
    3SFC w/ EF ($S=1$) & 87.63 & 48.50 & 76.73 & 76.00 & 57.92 & 24.20 & 33.34 & 4.39 & 6.14 & 78.27 & \cellcolor{orange!20}49.31 & \cellcolor{cyan!20}30.93 \\
    3SFC w/ EF ($S=5$) & 88.64 & 52.59 & 77.62 & 79.83 & 61.07 & 25.35 & 36.57 & 4.39 & 6.67 & 80.03 & \cellcolor{orange!20}51.27 & \cellcolor{cyan!20}31.45 \\
    3SFC w/ EF ($S=20$) & 89.02 & 54.97 & 79.31 & 81.86 & 62.08 & 26.63 & 37.61 & 4.52 & 7.24 & 81.74 & \cellcolor{orange!20}52.49 & \cellcolor{cyan!20}31.81 \\
    \midrule
    \midrule
    \multicolumn{13}{c}{20 Clients} \\
    \midrule
    3SFC w/ EF & 89.18 & 55.56 & 80.13 & 82.17 & 60.44 & 30.49 & 38.54 & 5.32 & 7.64 & 82.19 & \cellcolor{orange!20}53.16 & \cellcolor{cyan!20}31.36 \\
    3SFC w/o EF & 67.07 & 19.70 & 60.08 & 74.50 & 46.25 & 23.73 & 30.62 & 2.31 & 2.95 & 66.1& \cellcolor{orange!20}39.33 & \cellcolor{cyan!20}27.07 \\
    3SFC w/ EF ($2\times B$)   & 89.48 & 58.32 & 80.53 & 83.43 & 61.65 & 34.50 & 43.76 & 9.73 & 9.77 & 82.19 & \cellcolor{orange!20}55.33 & \cellcolor{cyan!20}29.97 \\
    3SFC w/ EF ($4\times B$)   & 89.63 & 59.61 & 80.70 & 83.72 & 61.87 & 36.54 & 45.08 & 11.18 & 10.31 & 82.39 & \cellcolor{orange!20}56.10 & \cellcolor{cyan!20}29.52 \\
    3SFC w/ EF ($K=1$) & 75.04 & 33.87 & 64.41 & 77.06 & 52.49 & 21.55 & 30.72 & 1.26 & 4.12 & 58.21 & \cellcolor{orange!20}41.87 & \cellcolor{cyan!20}27.69 \\
    3SFC w/ EF ($K=10$) & 90.63 & 61.27 & 82.89 & 82.94 & 60.49 & 30.95 & 41.50 & 4.83 & 8.31 & 85.92 & \cellcolor{orange!20}54.97 & \cellcolor{cyan!20}32.16 \\
    3SFC w/ EF ($S=1$) & 88.75 & 48.99 & 77.50 & 79.78 & 59.46 & 22.80 & 35.83 & 3.71 & 6.58 & 80.43 & \cellcolor{orange!20}50.38 & \cellcolor{cyan!20}31.84 \\
    3SFC w/ EF ($S=5$) & 89.21 & 55.70 & 79.77 & 82.64 & 60.66 & 30.46 & 37.50 & 4.75 & 7.57 & 82.06 & \cellcolor{orange!20}53.03 & \cellcolor{cyan!20}31.54 \\
    3SFC w/ EF ($S=20$) & 88.93 & 56.03 & 79.97 & 81.82 & 60.99 & 30.47 & 39.11 & 5.05 & 8.01 & 82.29 & \cellcolor{orange!20}53.26 & \cellcolor{cyan!20}31.27 \\
    \midrule
    \midrule
    \multicolumn{13}{c}{40 Clients} \\
    \midrule
    3SFC w/ EF  & 88.86 & 55.95 & 79.45 & 82.70 & 61.45 & 28.69 & 38.35 & 5.60 & 6.18 & 82.03 & \cellcolor{orange!20}52.92 & \cellcolor{cyan!20}31.68 \\
    3SFC w/o EF & 48.30 & 25.12 & 63.49 & 75.75 & 47.42 & 23.00  & 29.17 & 1.33 & 2.77 & 61.87 & \cellcolor{orange!20}39.82 & \cellcolor{cyan!20}27.22 \\
    3SFC w/ EF ($2\times B$)   & 89.32 & 58.76 & 80.27 & 83.74 & 61.32 & 37.47 & 44.81 & 10.41 & 7.99 & 82.06 & \cellcolor{orange!20}55.61 & \cellcolor{cyan!20}29.87 \\
    3SFC w/ EF ($4\times B$)   & 89.49 & 59.95 & 80.73 & 84.12 & 61.15 & 36.95 & 45.03 & 11.89 & 8.89 & 82.18 & \cellcolor{orange!20}56.03 & \cellcolor{cyan!20}29.62 \\
    3SFC w/ EF ($K=1$) & 69.56 & 31.29 & 65.47 & 77.52 & 52.43 & 22.54 & 30.24 & 1.27 & 3.17 & 56.87 & \cellcolor{orange!20}41.03 & \cellcolor{cyan!20}27.26 \\
    3SFC w/ EF ($K=10$) & 90.72 & 60.74 & 82.77 & 83.99 & 58.75 & 30.99 & 41.76 & 4.76 & 7.48 & 85.67 & \cellcolor{orange!20}54.76 & \cellcolor{cyan!20}32.33 \\
    3SFC w/ EF ($S=1$) & 88.34 & 50.85 & 78.02 & 81.14 & 58.91 & 26.38 & 36.24 & 4.95 & 5.28 & 79.91 & \cellcolor{orange!20}51.0& \cellcolor{cyan!20}31.56 \\
    3SFC w/ EF ($S=5$) & 88.97 & 55.31 & 79.49 & 83.03 & 61.82 & 27.59 & 38.12 & 5.42 & 6.10 & 81.9& \cellcolor{orange!20}52.77 & \cellcolor{cyan!20}31.87 \\
    3SFC w/ EF ($S=20$) & 88.99 & 55.96 & 79.28 & 83.19 & 61.49 & 31.65 & 38.83 & 6.32 & 6.24 & 81.97 & \cellcolor{orange!20}53.39 & \cellcolor{cyan!20}31.33 \\
    \midrule
    \midrule
    \multicolumn{13}{c}{60 Clients} \\
    \midrule
    3SFC w/ EF  & 89.05 & 48.86 & 80.28 & 82.55 & 59.65 & 29.61 & 39.84 & 7.02 & 6.93 & 82.0 & \cellcolor{orange!20}52.57 & \cellcolor{cyan!20}31.22 \\
    3SFC w/o EF & 70.92 & 4.48 & 64.33 & 75.75 & 45.31 & 21.67  & 30.34 & 1.01 & 2.46 & 61.94 & \cellcolor{orange!20}37.82 & \cellcolor{cyan!20}29.64 \\
    3SFC w/ EF ($2\times B$)   & 89.35 & 55.03 & 80.87 & 83.68 & 60.6 & 37.53 & 43.67 & 9.54 & 9.57 & 82.16 & \cellcolor{orange!20}55.2 & \cellcolor{cyan!20}29.82 \\
    3SFC w/ EF ($4\times B$)   & 89.71 & 56.8 & 81.26 & 83.76 & 60.8 & 37.62 & 44.4 & 10.81 & 10.17 & 82.21 & \cellcolor{orange!20}55.72 & \cellcolor{cyan!20}29.59 \\
    3SFC w/ EF ($K=1$) & 70.61 & 6.16 & 65.12 & 77.72 & 52.86 & 22.44 & 30.95 & 0.95 & 3.84 & 56.99 & \cellcolor{orange!20}38.76 & \cellcolor{cyan!20}29.43 \\
    3SFC w/ EF ($K=10$) & 90.64 & 56.96 & 82.93 & 83.62 & 58.28 & 32.81 & 43.16 & 5.05 & 7.8 & 85.6 & \cellcolor{orange!20}54.68 & \cellcolor{cyan!20}31.92 \\
    3SFC w/ EF ($S=1$) & 88.36 & 32.11 & 78.49 & 80.97 & 58.24 & 26.62 & 37.11 & 5.99 & 5.87 & 79.34 & \cellcolor{orange!20}49.31 & \cellcolor{cyan!20}31.76 \\
    3SFC w/ EF ($S=5$) & 89.06 & 47.7 & 80.14 & 82.27 & 59.59 & 30.1 & 39.86 & 7.16 & 6.46 & 81.96 & \cellcolor{orange!20}52.43 & \cellcolor{cyan!20}31.21 \\
    3SFC w/ EF ($S=20$) & 89.11 & 48.94 & 80.15 & 82.73 & 60.59 & 29.99 & 39.78 & 7.22 & 6.86 & 82.07 & \cellcolor{orange!20}52.74 & \cellcolor{cyan!20}31.22 \\
    \bottomrule
    \bottomrule
    \end{tabular}%
    }
  \caption{The hyper-parameter analysis with different parameters of 3SFC (\textit{i.e.}, with/without EF, communication budgets $B$, local iteration $K$, number of 3SFC iteration $S$). \colorbox{orange!20}{Avg} and \colorbox{cyan!20}{Std}: the average results and the standard deviation. The configuration for the Base is 1 $\times B$, $K$ = 5 and $S$ = 10. From the table, it is clear that enabling EF in 3SFC has an important role in helping models converge, and validating its effectiveness. Moreover, Increasing $B$, $K$ or $S$ can both further boost the convergence rate of the training.}
  \label{tab:hyper-analysis-3sfc}%
\end{table*}%
\begin{figure}
     \centering
     \begin{subfigure}[tb]{0.20\textwidth}
         \centering
         \includegraphics[width=\textwidth]{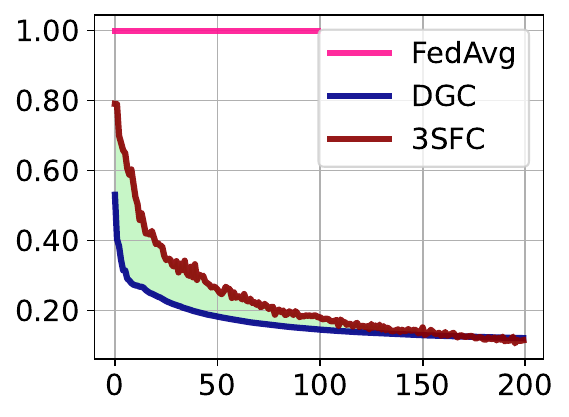}
         \caption{MLP on EMNIST.}
     \end{subfigure}
     \begin{subfigure}[tb]{0.20\textwidth}
         \centering
         \includegraphics[width=\textwidth]{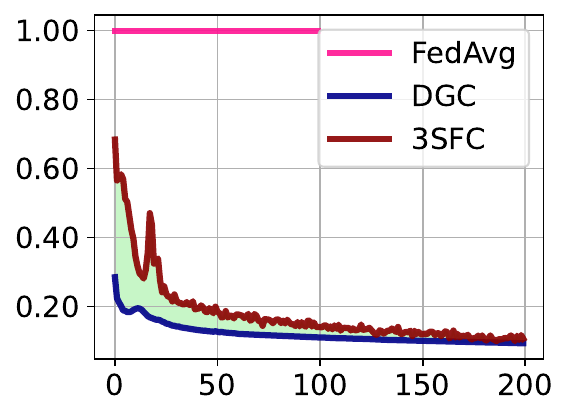}
         \caption{MnistNet on FMNIST.}
     \end{subfigure}
     \\
     \begin{subfigure}[tb]{0.20\textwidth}
         \centering
         \includegraphics[width=\textwidth]{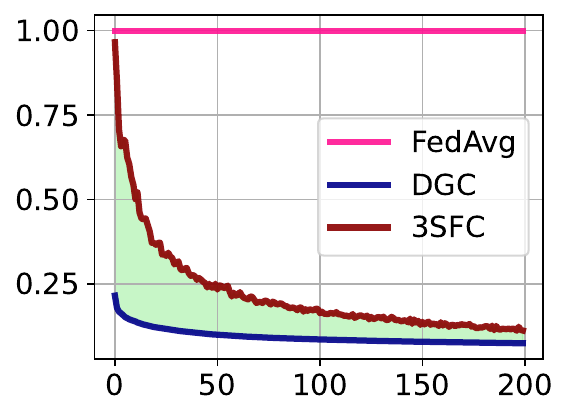}
         \caption{RegNet on Cifar10.}
     \end{subfigure}
     \begin{subfigure}[tb]{0.20\textwidth}
         \centering
         \includegraphics[width=\textwidth]{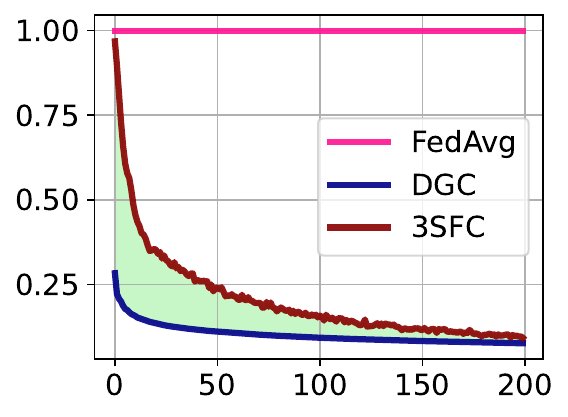}
         \caption{RegNet on Cifar100.}
     \end{subfigure}
     \\
     \begin{subfigure}[tb]{0.20\textwidth}
         \centering
         \includegraphics[width=\textwidth]{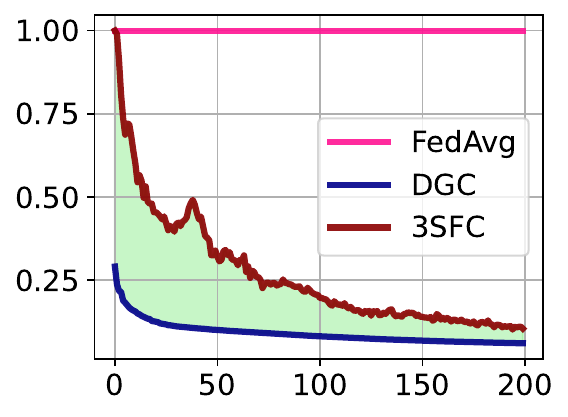}
         \caption{ResNet on Cifar10.}
     \end{subfigure}
     \begin{subfigure}[tb]{0.20\textwidth}
         \centering
         \includegraphics[width=\textwidth]{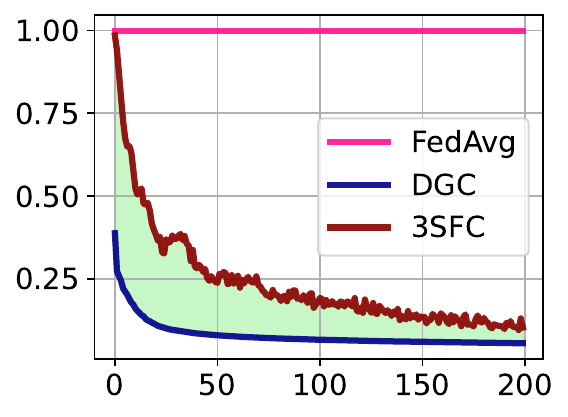}
         \caption{ResNet on Cifar100.}
     \end{subfigure}
        \caption{Compression efficiency comparisons. 3SFC owns significantly higher compression efficiency compared to DGC, validating its effectiveness.}
        \label{fig:compression-efficiency}
\end{figure}

\subsection{Further Comparison Analysis}
\label{sec:3sfc-stc-comp}
In Table~\ref{tab:accuracy-compare}, the best performing quantification-based method, STC, sometimes surpasses 3SFC and E-3SFC in terms of test accuracy due to its low compression ratio.
To further evaluate our proposed methods compared to STC for fairness, we gradually increase the communication budget in 3SFC and compare both their compression ratio and test accuracy simultaneously, as shown in Table~\ref{tab:3sfc-stc-comp}.
Here, we only compare STC with 3SFC because: 1) STC and 3SFC are compressors while E-3SFC is an algorithm; 2) E-3SFC includes 3SFC, and from Table~\ref{tab:accuracy-compare}, E-3SFC outperforms 3SFC under the same compression ratio across all settings.
As the table suggests, 3SFC can achieve comparable or even better test accuracy while saving a significant amount of communication traffic. For example, when training ResNet on Cifar10 with 10 clients, 3SFC reports a comparable final accuracy (39.54 v.s. 40.09) while communicating 27.9$\times$ fewer data. On the other hand, 3SFC reaches considerably better performance for RegNet on Cifar100 (9.46 v.s. 4.16) with a 6.0$\times$ better compression ratio.

\subsection{Compression Efficiency}
To study why 3SFC achieves a faster convergence rate, we restrain the compression ratio of 3SFC and DGC to be the same, and visualize the compression efficiency $\mathcal{E}(t)$ of 3SFC, DGC and FedAvg. Here, the compression efficiency stands for how much information the compressed data carry compared to the uncompressed data, defined by:
\begin{equation}
    \mathcal{E}(t)=\cos(Decompressed,Uncompressed).
\end{equation}

The visualization is shown in Figure~\ref{fig:compression-efficiency}. 
In Figure~\ref{fig:compression-efficiency}, FedAvg has a constant compression efficiency of 1.0, as FedAvg does not compress the data at all. 
Hence, FedAvg is served as a reference here. Meanwhile, it is clear from the figure that with the same compression ratio, 3SFC achieves higher compression efficiency for every communication round (\textit{i.e.}, the green area), meaning that the compression error of 3SFC is lower at every update step of the global model, contributing to the faster convergence rate of the training. Moreover, as error feedback is incorporated into both DGC and 3SFC, the compression error of each communication round will be accumulated into $\mathbf{g}_i^t$ forever. Consequently, the compression efficiency for both DGC and 3SFC decreases gradually as the training progresses.

\begin{table*}[htbp]
  \centering
  \resizebox{0.78\linewidth}{!}{%
    \begin{tabular}{lrrrrrrrrrrr}
    \toprule
    \toprule
    \multicolumn{1}{c}{Budget} & \multicolumn{1}{c}{MNIST} & \multicolumn{1}{c}{EMNIST} & \multicolumn{2}{c}{FMNIST} & \multicolumn{3}{c}{Cifar10} & \multicolumn{2}{c}{Cifar100} & \multicolumn{2}{c}{Statistics} \\
\cmidrule{4-12}    \multicolumn{1}{c}{Scheduler} & \multicolumn{1}{c}{MLP} & \multicolumn{1}{c}{MLP} & \multicolumn{1}{c}{MLP} & \multicolumn{1}{c}{MnistNet} & \multicolumn{1}{c}{ConvNet} & \multicolumn{1}{c}{ResNet} & \multicolumn{1}{c}{RegNet} & \multicolumn{1}{c}{ResNet} & \multicolumn{1}{c}{RegNet} & \multicolumn{1}{c}{\cellcolor{orange!20}Avg} & \multicolumn{1}{c}{\cellcolor{cyan!20}Std} \\
    \midrule
    \midrule
    \multicolumn{12}{c}{10 Clients} \\
    \midrule
    Constant & 89.58 & 59.55 & 80.63 & 84.30 & 62.41 & 39.54 & 43.41 & 9.89 & 9.52 & \cellcolor{orange!20}53.20 & \cellcolor{cyan!20}30.07 \\
    Linear & 89.56 & \underline{59.74} & 80.91 & 83.60 & 61.83 & \underline{39.74} & 43.87 & 9.32 & \underline{9.68} & \cellcolor{orange!20}53.13 & \cellcolor{cyan!20}30.04 \\
    Cosine & \underline{89.63} & 59.56 & \underline{80.99} & \underline{84.34} & \textbf{62.41} & 39.50 & \underline{44.19} & \textbf{10.09} & 9.66 & \cellcolor{orange!20}53.37 & \cellcolor{cyan!20}30.04 \\
    Eq.~\ref{eq:objective_budget_scheduler_sim}  & \textbf{89.71} & \textbf{59.80} & \textbf{81.04} & \textbf{84.52} & \underline{62.38} & \textbf{39.96} & \textbf{44.48} & \underline{10.08} & \textbf{9.82} & \cellcolor{orange!20}53.53 & \cellcolor{cyan!20}30.02 \\
    \midrule
    \midrule
    \multicolumn{12}{c}{20 Clients} \\
    \midrule
    Constant & 89.63 & 59.61 & 80.70 & 83.72 & 61.87 & 36.54 & 45.08 & 11.18 & \underline{10.31} & \cellcolor{orange!20}53.18 & \cellcolor{cyan!20}29.75 \\
    Linear & 89.68 & 59.68 & 81.03 & 83.82 & 61.11 & 36.58 & 44.67 & 10.65 & 9.87 & \cellcolor{orange!20}53.01 & \cellcolor{cyan!20}29.97 \\
    Cosine & \underline{89.68} & \textbf{59.83} & \underline{81.09} & \underline{83.84} & \underline{61.99} & \textbf{37.09} & \underline{45.12} & \underline{11.28} & 10.28 & \cellcolor{orange!20}53.35 & \cellcolor{cyan!20}29.77 \\
    Eq.~\ref{eq:objective_budget_scheduler_sim}  & \textbf{89.72} & \underline{59.75} & \textbf{81.13} & \textbf{84.25} & \textbf{62.08} & \underline{36.73} & \textbf{45.21} & \textbf{11.35} & \textbf{10.39} & \cellcolor{orange!20}53.40 & \cellcolor{cyan!20}29.83 \\
    \midrule
    \midrule
    \multicolumn{12}{c}{40 Clients} \\
    \midrule
    Constant & 89.49 & 59.95 & \underline{80.73} & \underline{84.12} & 61.15 & 36.95 & \underline{45.03} & 11.89 & 08.89 & \cellcolor{orange!20}53.13 & \cellcolor{cyan!20}29.88 \\
    Linear & \underline{89.54} & 60.02 & 80.64 & 84.03 & 60.65 & 36.54 & 44.30 & 11.04 & 8.75 & \cellcolor{orange!20}52.83 & \cellcolor{cyan!20}30.07 \\
    Cosine & 89.53 & \underline{60.03} & 80.53 & 84.10 & \underline{61.20} & \underline{36.99} & 45.01 & \underline{11.89} & \underline{8.98} & \cellcolor{orange!20}53.14 & \cellcolor{cyan!20}29.84 \\
    Eq.~\ref{eq:objective_budget_scheduler_sim}  & \textbf{89.56} & \textbf{60.05} & \textbf{80.77} & \textbf{84.23} & \textbf{61.29} & \textbf{37.19} & \textbf{45.29} & \textbf{12.02} & \textbf{9.05} & \cellcolor{orange!20}53.27 & \cellcolor{cyan!20}29.83 \\
    \bottomrule
    \bottomrule
    \end{tabular}%
  }
  \caption{Comparisons of test accuracy with different budget schedulers under the relaxed communication budget constraint. We mark the best and second best performance by \textbf{bold} and \underline{underline}. \colorbox{orange!20}{Avg} and \colorbox{cyan!20}{Std}: the average results and the standard deviation. 
  Our proposed budget scheduler solving Eq.~\ref{eq:objective_budget_scheduler_sim} outperforms no scheduler (\textit{i.e.}, Constant).}
  \label{tab:budget-scheduler}%
\end{table*}%

\subsection{Ablation Study and Hyper-parameter Analysis}
In this section, we report ablation studies on the E-3SFC's sub-components and the hyper-parameter analysis of the two sub-components, \textit{i.e}, the 3SFC and the budget scheduler. 

Table~\ref{tab:ablation-e3sfc} shows the ablation studies on the Budget Scheduler (BS), Double-way Compression (DWC) and EF of E-3SFC.
One can see that all sub-components improve the test performance.
Specifically, disabling EF in E-3SFC drastically degrades the model performance after training in all experiments, validating the effectiveness of EF.
Moreover, the test performance can also be considerably increased by either enabling double-way compression or dynamically scheduling the communication budget with the same compression ratio.
These experimental results suggest that all sub-components contribute to enhancing communication efficiency of E-3SFC.

To further analyze impacts of hyper-parameters in 3SFC, hyper-parameter analyses on the communication budget $B$, local iteration $K$, iteration $S$ of 3SFC are conducted and showed in Table~\ref{tab:hyper-analysis-3sfc}. 
As it can be seen, when increasing $B$, the test accuracy of the model increases as well, as more data can be transferred at each communication round to effectively reduce the compression error. On the other hand, by decreasing the local iteration $K$ from 5 to 1, the test accuracy is reduced significantly since the model has been optimized much less. Contrarily, the test accuracy boosts up when $K$ is set to 10. Moreover, increasing $S$ from $1$ to $5$ also substantially helps the model converge since synthetic features can be optimized better. However, setting $S$ to $20$ has little or even no effect on the final model performance, suggesting there is a marginal effect for $S$ and its value should be selected based on the downstream task. Consequently, increasing the communication budget $B$ is the most significant way to further boost the convergence rate of the model using 3SFC. For example, by increasing the compression budget from $B$ to $2\times B$ for ResNet trained on Cifar100 with 40 clients, the convergence rate of the model gets doubled and the final test accuracy after 200 epochs of training also increases from 5.60 to 10.41. However, when the communication budget is strictly limited, the convergence rate of 3SFC can be improved as well by setting a larger local iteration $K$ or 3SFC iteration $S$.

Finally, under the relaxed communication budget constraint, we conduct analyses on our proposed budget scheduler and compare it with no budget scheduler and other vanilla schedulers like the linear scheduler and the cosine scheduler in Table~\ref{tab:budget-scheduler}.
Specifically, the linear scheduler adjusts the communication budget by $H(B, t, i) = (1 - B) ((t + iT/N) \mod T) / T  + B$ and the cosine scheduler adjusts the communication budget by $H(B, t, i) = (B - 1) (1 - \cos(((t + iT/N) \mod T)\pi / T) / 2 + 1$.
For our proposed budget scheduler, to obtain its closed-form solution, Eq.~\ref{eq:objective_budget_scheduler_sim} is optimized using Sequential Least Squares Programming Algorithm (SLSQP) in advance with $\tau=3.0$.
From the table, it can be noticed that models trained using our proposed budget scheduler mostly achieve better performance compared to other budget schedulers, suggesting our proposed budget scheduler successfully takes better advantage of limited communication budgets.

\section{Conclusion}
\label{sec:conclusion}
In this paper, we propose a systematic gradient compression algorithm termed E-3SFC, which consists of a gradient compressor (3SFC), double-way compression, and a communication budget scheduler.
The proposed compressor 3SFC utilizes training priors to effectively reduce compression errors.
In detail, 3SFC compresses the data using a similarity-based objective function with single-step simulation, thus saving both computation and storage resources and maintaining the robustness of the algorithm. 
Moreover, error feedback is employed to further minimize the compression error.
Additionally, to further improve the communication efficiency, a double-way compression algorithm is introduced to compress the updated global model in the model downloading phase and to employ a novel budget scheduler under the relaxed communication constraint for enhanced flexibility in dynamic communication budget adjustment.
Detailed theoretical analyses are provided to guarantee that 3SFC has a linear and sub-linear speed-up convergence rate with aggregation noise under the strongly convex and the non-convex settings, respectively
Extensive experiments across various datasets and models demonstrate that E-3SFC outperforms existing methods, achieving up to 13.4\% improvement in performance while reducing communication costs by 111.6 times.
Other analyses also confirm the effectiveness of our proposals in this paper.

Future work will focus on further optimizing the compression algorithm to enhance its scalability and robustness in more diverse and dynamic network environments.
We also plan to explore the integration of E-3SFC with other advanced FL techniques, such as personalized FL (\textit{e.g.}, pFedMe~\cite{t2020personalized}, PRIOR~\cite{shi2023prior}, FCCA~\cite{zhou2024federated}, etc.) and error feedback strategies (\textit{e.g.}, SAEF~\cite{xu2021step}, DEF~\cite{xu2022detached}, etc.), to further improve the efficiency and effectiveness of the learning process.
Additionally, investigating the applicability of our method to other FL paradigms, such as federated reinforcement learning~\cite{qi2021federated}, could open new avenues for research and applications.

\section*{Acknowledgments}
This work was supported by National Natural Science Foundation of China (Grant 62306198) and Natural Science Foundation of Sichuan Province (Grant 2024NSFSC1468).

\bibliographystyle{IEEEtran}

\section{Biography Section}
\vspace{-30pt}
\begin{IEEEbiography}[{\includegraphics[width=1in,height=1.25in,clip,keepaspectratio]{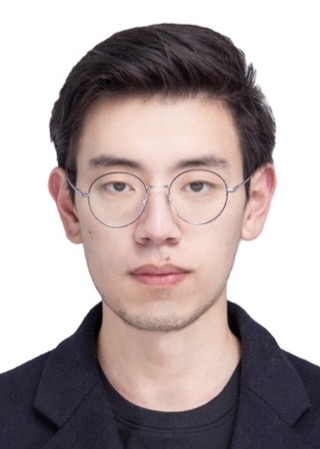}}]{Yuhao Zhou}
is currently working toward a Ph.D. degree in computer science with the College of Computer Science, Sichuan University, China. He received the B.E. degree in College of Computer Science, Sichuan University, China, in 2022. His main research interests include distributed machine learning, federated learning, and optimization. His researches have been published in several journals and conferences such as IEEE TPDS, ICCV, NeurIPS, ICASSP, IEEE TETCI, ICONIP, etc.
\end{IEEEbiography}
\vspace{-15pt}
\begin{IEEEbiography}[{\includegraphics[width=1in,height=1.25in,clip,keepaspectratio]{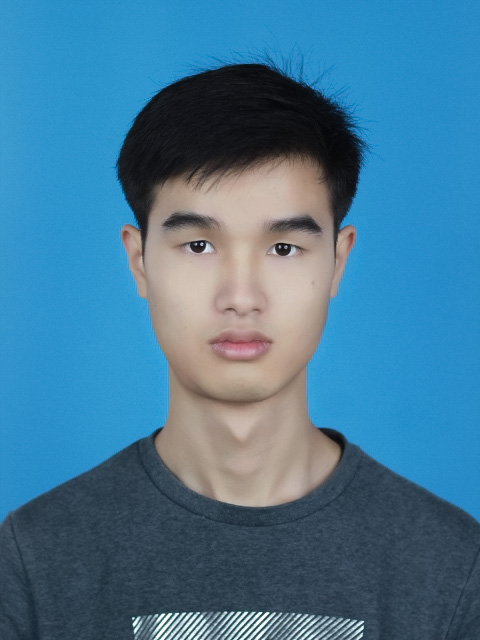}}]{Yuxin Tian}
is currently pursuing a Ph.D. degree in Computer Science for machine intelligence. He received the B.E. degree in College of Computer Science, Sichuan University, Chengdu, China, in 2020. His research interests include multimodal learning, federated learning, learning with noisy labels and so on. His research has been published on IEEE T SYST MAN CY-S, IEEE T-NNLS, ICML, ICME, ICASSP, etc.
\end{IEEEbiography}
\vspace{-15pt}
\begin{IEEEbiography}[{\includegraphics[width=1in,height=1.25in,clip,keepaspectratio]{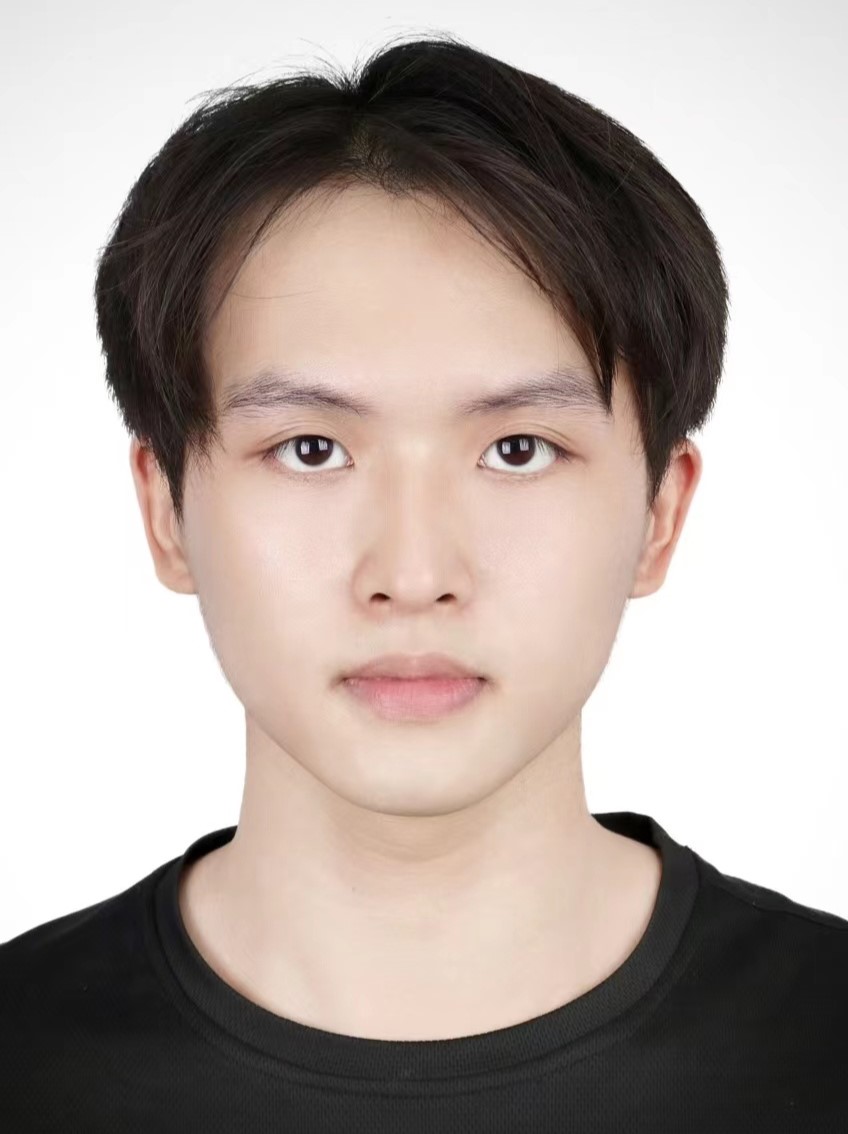}}]{Minjia Shi}
is currently pursuing a master degree in artificial intelligent. He received the B.Edegree in College of Computer Science Sichuan University, Chengdu, in 2019. His current interests mainly on transfer learning, federated learning and optimization. His researches have been published in several journals and conferences such as ICCV, NeuraIPS, IEEE TETCI, etc.
\end{IEEEbiography}
\vspace{-15pt}
\begin{IEEEbiography}[{\includegraphics[width=1in,height=1.25in,clip,keepaspectratio]{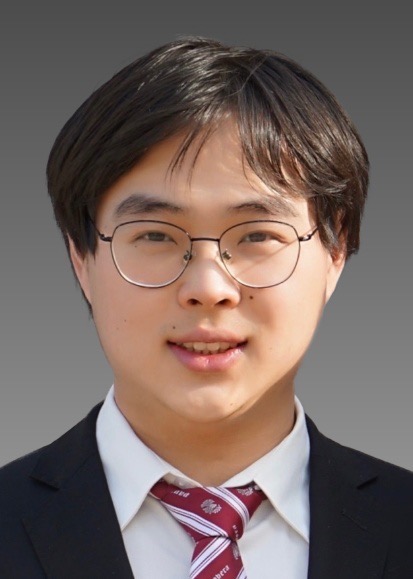}}]{Yuanxi Li}
received the Bachelor of Science Degree in Mathematics \& Computer Science at University of Illinois Urbana-Champaign in 2022. His work has been published in several journals and conferences, including ICCV, EMNLP, and ICASSP.
\end{IEEEbiography}
\vspace{-15pt}
\begin{IEEEbiography}[{\includegraphics[width=1in,height=1.25in,clip,keepaspectratio]{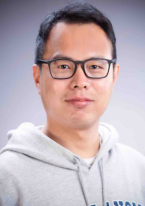}}]{Yanan Sun} received the Ph.D. degree in computer science from Sichuan University, Chengdu, China, in 2017. He is currently a Professor with the College of Computer Science, Sichuan University. His research interests include evolutionary computation, neural networks, and their applications on neural architecture search. He designed the indicator of “GPU Day”, which has been widely used among the community of neural architecture search. He was ranked as World's Top 2\% Scientists 2021, collectively released by Stanford University and Springer. He is an Associate Editor for IEEE Transactions on Evolutionary Computation and IEEE Transactions on Neural Networks and Learning Systems.
\end{IEEEbiography}
\vspace{-15pt}
\begin{IEEEbiography}[{\includegraphics[width=1in,height=1.25in,clip,keepaspectratio]{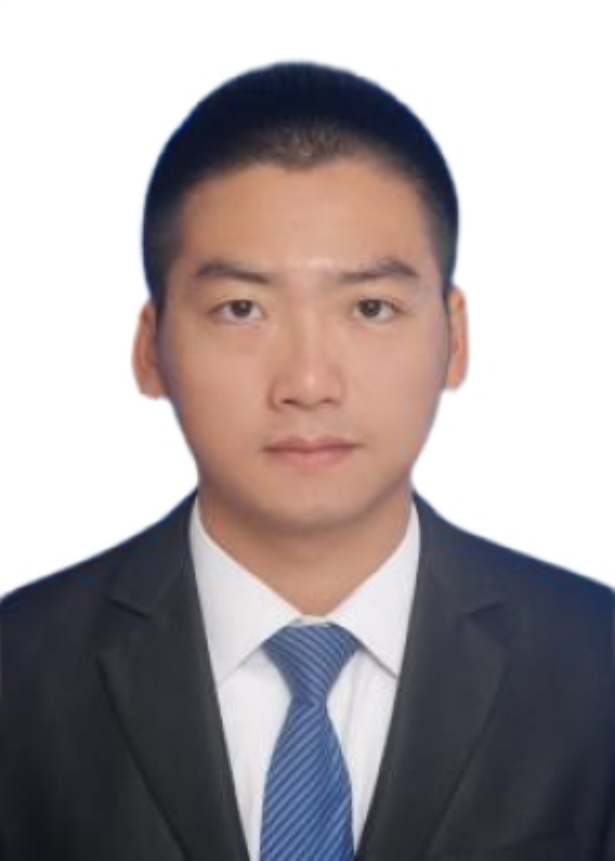}}]{Qing Ye}
received his Ph.D. degree in Computer Science from Sichuan University, Chengdu, China in 2022. He is currently a lecturer at the School of Computer Science, Sichuan University, Chengdu, China. His main research interests include distributed machine learning, deep learning and neural architecture search.
\end{IEEEbiography}
\vspace{-15pt}
\begin{IEEEbiography}[{\includegraphics[width=1in,height=1.25in,clip,keepaspectratio]{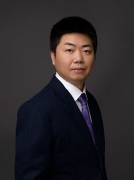}}]{Jiancheng Lv}
is the professor and dean of the School of Computer Science (School of Software) of Sichuan University. He received the Ph.D. degree in Computer Science at the University of Electronic Science and Technology of China since 2006. His researching interests include subspace learning in neural network, natural language generation, and computer vision. His researches have been published in several journals and conferences such as IEEE TNNLS, IEEE TFS, IEEE TSMC, IEEE TCYB, NeurIPS, ICML, ACL, CVPR, AAAI, IJCAI, etc.
\end{IEEEbiography}

 

% \end{IEEEbiography}



\vfill
\newpage

{
\appendices

\section*{Proof of Lemma~\ref{lemma:bounded-local-shift}}
\label{appendix:proof_lemma_1}
\begin{proof}
\begingroup
\allowdisplaybreaks
\begin{align}\label{eq:lemma_1_1}
&\Delta^{t_a + K} =  \Delta^{t_a} - \underbrace{2 \langle w^{t_a} - w^*, \sum_{k=0}^{K-1} g_i^{t_a + k} \rangle}_{A} + \underbrace{||\sum_{k=0}^{K-1} g_i^{t_a + k}||^2}_{B}
\end{align}
\endgroup

For the term $A$, with $\langle w^{t_a} - w^*, \nabla F(w^{t_a}) \rangle \geq F(w^{t_a}) - F(w^*) + \frac{\mu_F}{2} ||w^{t_a} - w^*|| ^ 2$, let $\tilde{\eta} = \sum_{k=0}^{K-1} K \eta$, we have: 
\begin{align}
    A &= -2 \tilde{\eta} \langle w^{t_a} - w^*, \frac{\sum_{k=0}^{K-1} g_i^{t_a + k}}{\tilde{\eta}} - \nabla F (w^{t_a}) \rangle \nonumber \\ 
    &~~~- 2 \langle w^{t_a} - w^*, \tilde{\eta} \rangle \\
    &\leq \frac{\mu_F \tilde{\eta}}{2} ||w^{t_a} - w^*||^2 + \frac{2 \tilde{\eta}}{\mu_F} || \sum_{k=0}^{K-1} \frac{g_i^{t_a + k}}{\tilde{\eta}}||^2 \nonumber \\
    &~~~- 2 \tilde{\eta} \langle w^{t_a} - w^*, \nabla F (w^{t_a}) \rangle \\
    & \leq -\frac{\tilde{\eta} \mu_F}{2} ||w^{t_a} - w^*||^2 + \frac{2}{\mu_F \tilde{\eta}} || \sum_{k=0}^{K-1} g_i^{t_a + k} - \tilde{\eta} \nabla F(w^{t_a}) ||^2 \nonumber \\
    &~~~- 2\tilde{\eta}(F(w^{t_a}) - F(w^*).
\label{eq:lemma_1_2}
\end{align}

For the term $B$, with Jensen's inequation, we have:
\begin{align}
    \mathbb{E}B &= \mathbb{E} \tilde{\eta}^2 ||\frac{\sum_{k=0}^{K-1} g_i^{t_a + k}}{\tilde{\eta}}||^2 \\
    &= \mathbb{E} \tilde{\eta}^2 ||\frac{1}{M\tilde{\eta}} \sum_{i, k=0}^{M^t, K-1} g_i^{t_a+k}||^2 \\
    &\leq 3\mathbb{E} \tilde{\eta}^2 (||\frac{1}{M\tilde{\eta}} \sum_{i, k=0}^{M^t, K-1} g_i^{t_a+k} - \nabla F(w^{t_a}) ||^2 \nonumber \\
    &~~~+ ||\frac{1}{M} \sum_{i}^{M^t} \nabla F(w^{t_a}) - \nabla F(w^{t_a})||^2 + ||\nabla F(w^{t_a})||^2).
\label{eq:lemma_1_3}
\end{align}

Finally, Taking Inequation~\ref{eq:lemma_1_2} and Inequation~\ref{eq:lemma_1_3} into Equation~\ref{eq:lemma_1_1}, Inequation~\ref{lemma_1_4} can be derived.
\begin{align}
    &~~~\mathbb{E}||w^{t_a + K} - w^*||^2 \nonumber \\
    &\leq (1 - \frac{\tilde{\eta} \mu_F}{2}) \mathbb{E} ||w^{t_a} - w^*||^2 \nonumber \\
    &~~~+ \frac{(\frac{2}{\mu_F \tilde{\eta}} + 3)K}{N} \sum_{i=0,k=0}^{N,K-1} \mathbb||g_i^{t_a + k} - \eta^k \nabla F(w^{t_a})||^2 \nonumber \\
    &~~~- 2\tilde{\eta} \mathbb{E} (F(w^{t_a}) - F(w^*)) + 3\tilde{\eta}^2 \mathbb{E} ||\nabla F(w^{t_a})||^2  \nonumber \\
    &~~~+ 3\tilde{\eta}^2 \mathbb{E} ||\frac{1}{M} \sum_{i=0}^{M^t} \nabla F(w^{t_a}) - \nabla F(w^{t_a})||^2 \\ 
    &\leq (1 - \frac{\tilde{\eta} \mu_F}{2}) \mathbb{E} ||w^{t_a} - w^*||^2 \nonumber \\
    &~~~+ \frac{(\frac{2}{\mu_F \tilde{\eta}} + 3)K}{N} \sum_{i=0,k=0}^{N,K-1} \mathbb||g_i^{t_a + k} - \eta^k \nabla F(w^{t_a})||^2 \nonumber \\
    &~~~+ (6L\tilde{\eta}^2 - 2\tilde{\eta}) \mathbb{E}(F(w^{t_a}) - F(w^*)) \nonumber \\
    &~~~+ 3\tilde{\eta}^2 \mathbb{E} ||\frac{1}{M} \sum_{i}^{M^t} \nabla F(w^{t_a}) - \nabla F(w^{t_a})||^2,
\label{lemma_1_4}
\end{align}
and the Lemma~\ref{lemma:bounded-local-shift} is proved.
\end{proof}

\section*{Proof of Lemma~\ref{lemma:bounded_local_approximation}}
\label{appendix:proof_lemma_2}
\begin{proof}
Let $\dot{\delta}:=3(\kappa_S^2+\sigma_F^2)$ and $\tilde{\eta} \leq \frac{1}{6L}$, we have $3 \eta^2 L \leq \frac{1}{K^2} \leq \frac{1}{2K}$. Therefore, inequation~\ref{lemma_2_1} can be obtained.
\begin{align}
    &~~~\mathbb{E}||g_i^{t_a+k} - \eta \nabla F(w^{t_a})||^2 \nonumber \\
    &\leq 3 \mathbb{E}(||g_i^{t_a+k} - \eta \nabla \tilde{F}(w^{t_a + k})||^2 \nonumber \\
    &~~~+ \eta^2\nabla \tilde{F}(w^{t_a + k}) - \nabla F(w^{t_a + k})||^2 \nonumber \\
    &~~~+ \eta^2\nabla F(w^{t_a + k}) - \nabla F(w^{t_a})||^2 ) \\
    &\leq 3 \eta^2(\kappa_S^2 + \sigma_F^2 + L\mathbb{E}||w^{t_a + k} - w^{t_a}||^2) \\
    &\leq 3 \eta^2 L \underbrace{\mathbb{E}||w^{t_a + k} - w^{t_a}||^2}_{C} + \eta^2 \dot{\delta}.
\label{lemma_2_1}
\end{align}

By expanding $w^{t_a + k}$, We have:
\begin{align}
    C &\leq ||w_i^{t_a + k - 1} - w^{t_a} - g_i^{t_a + k - 1}||^2 \\
    &\leq 2(1 + \frac{1}{2K}) ||w_i^{t_a + k - 1} - w^{t_a}||^2 \nonumber \\
    &~~~+ 2(1 + 2K) \eta^2 ||\nabla F(w^{t_a})||^2 \nonumber \\
    &~~~+ 2 ||g_i^{t_a + k - 1} - \eta \nabla F(w^{t_a})||^2 \\
    &\leq 2(1 + \frac{1}{2K} + 3 \eta^2 L) ||w_i^{t_a + k - 1} - w^{t_a}||^2 \nonumber \\
    &~~~+ 2 \eta^2 (1 + 2K) ||\nabla F(w^{t_a})||^2 + 2 \eta^2 \dot{\delta} \\
    &\leq 2(1 + \frac{1}{K}) ||w_i^{t_a + k - 1} - w^{t_a}||^2 \nonumber \\
    &~~~+ 2 \eta^2 (1 + 2K) ||\nabla F(w^{t_a})||^2 + 2 \eta^2 \dot{\delta} \\
    &\leq \eta^2 ((1 + 2K)||\nabla F(w^{t_a})||^2 + \dot{\delta}) \sum_{\tilde{k} = 0}^{k} 2^{\tilde{k} + 1}(1 + \frac{1}{K})^{\tilde{k}} \\
    &\leq \eta^2 ((1 + 2K)||\nabla F(w^{t_a})||^2 + \dot{\delta}) e 2^{K+1}.
\label{lemma_2_2}
\end{align}

Thus, by taking inequation~\ref{lemma_2_2} into inequation~\ref{lemma_2_1}, we have:
\begin{align}
    &~~~\mathbb{E}||g_i^{t_a+k} - \eta \nabla F(w^{t_a})||^2 \nonumber \\
    &\leq 3\eta^4 L e 2^{K+1} ((1 + 2K)||\nabla F(w^{t_a})||^2 + \dot{\delta}) + \eta^2 \dot{\delta},
\end{align}
and the Lemma~\ref{lemma:bounded_local_approximation} is proved.
\end{proof}

\section*{Proof of Theorem~\ref{theorem:theorem_1}}
\begin{proof}
Lemma~\ref{lemma:bounded-local-shift} can be further expanded into:
\begin{align}
    &~~~\mathbb{E}||w^{t_a + K} - w^*||^2 \nonumber \\
    &\leq (1 - \frac{\tilde{\eta} \mu_F}{2}) \mathbb{E} ||w^{t_a} - w^*||^2 \nonumber \\
    &~~~+ \frac{(\frac{2}{\mu_F \tilde{\eta}} + 3)K}{N} \sum_{i=0,k=0}^{N,K-1} \mathbb||g_i^{t_a + k} - \eta^k \nabla F(w^{t_a})||^2 \nonumber \\
    &~~~+ (6L\tilde{\eta}^2 - 2\tilde{\eta}) \mathbb{E}(F(w^{t_a}) - F(w^*)) \nonumber \\
    &~~~+ 3\tilde{\eta}^2 \frac{N/M - 1}{N - 1} \mathbb{E} ||\nabla F(w^{t_a}) - \nabla F(w^{t_a})||^2 \\
    &\leq (1 - \frac{\tilde{\eta} \mu_F}{2}) \mathbb{E} ||w^{t_a} - w^*||^2 \nonumber \\
    &~~~+ \frac{(\frac{2}{\mu_F \tilde{\eta}} + 3)K}{N} \sum_{i=0,k=0}^{N,K-1} \mathbb||g_i^{t_a + k} - \eta^k \nabla F(w^{t_a})||^2 \nonumber \\
    &~~~+ (6L\tilde{\eta}^2 - 2\tilde{\eta}) \mathbb{E}(F(w^{t_a}) - F(w^*)) \nonumber \\
    &~~~+ 3\tilde{\eta}^2 \frac{N/M - 1}{N - 1} \mathbb{E} ||\nabla F(w^{t_a}) - \nabla F(w^*)||^2 \nonumber \\
    &~~~+ 3\tilde{\eta}^2 \frac{N/M - 1}{N - 1} \mathbb{E}|| \nabla F(w^*)||^2 \\ 
    &\leq (1 - \frac{\tilde{\eta} \mu_F}{2}) \mathbb{E} ||w^{t_a} - w^*||^2 \nonumber \\
    &~~~+ \frac{(\frac{2}{\mu_F \tilde{\eta}} + 3)K}{N} \sum_{i=0,k=0}^{N,K-1} \mathbb||g_i^{t_a + k} - \eta^k \nabla F(w^{t_a})||^2 \nonumber \\
    &~~~+ (6L\tilde{\eta}^2 - 2\tilde{\eta}) \mathbb{E}(F(w^{t_a}) - F(w^*)) \nonumber \\
    &~~~+ 3\tilde{\eta}^2 \frac{N/M - 1}{N - 1} (2L(F(w^{t_a}) - F(w^*)) + \sigma_*^2).
\label{theorem_1_1}
\end{align}

Taking Lemma~\ref{lemma:bounded_local_approximation} into inequation~\ref{theorem_1_1}, we have:
\begin{align}
    &~~~\mathbb{E}||w^{t_a + K} - w^*||^2 \nonumber \\
    &\leq (1 - \frac{\tilde{\eta} \mu_F}{2}) \mathbb{E} ||w^{t_a} - w^*||^2 \nonumber \\
    &~~~+ (6L \tilde{\eta}^2(1 + \frac{N/M - 1}{N - 1}) - 2\tilde{\eta}) \mathbb{E}(F(w^{t_a}) - F(w^*)) \nonumber \\
    &~~~+ 3||\tilde{\eta}||^2 \frac{N/M - 1}{N - 1} \sigma_*^2 + \frac{(\frac{2}{\mu_F \tilde{\eta}} + 3)K}{N} \sum_{i=0,k=0}^{N,K-1} D,
\end{align}
where $D = \mathbb{E}(3 \tilde{\eta}^2 L (\tilde{\eta}^2((1 + 2K)||\nabla F(w^{t_a})||^2 + \dot{\delta})e 2^{K+1}) + \tilde{\eta}^2 \dot{\delta}$. It is trivial that:
\begin{align}
    \sum_{i=0,k=0}^{N,K-1}D &\leq \sum_{i=0,k=0}^{N,K-1}\mathbb{E}(3 {\eta^k}^4 L e 2^{K + 1} (4 L (1 + 2K) \nonumber \\
    &~~~(F(w^{t_a}) - F(w^*)) + 2\sigma_*^2 + \dot{\delta}) + {\eta^k}^2 \dot{\delta}).
\end{align}

Thus, by choosing an proper $\tilde{\eta} \leq \min\{\hat{\eta}, \frac{2}{\mu_F}, \frac{1}{L}\}$, where $\hat{\eta} := \frac{\mu_F K}{3L[4\mu_F R + e 2^{K+6}(1/K+2)]}$, we have:
\begin{align}
    &~~~\mathbb{E}||w^{t_a + K} - w^*||^2 \nonumber \\
    &\leq (1 - \frac{\tilde{\eta} \mu_F}{2}) \mathbb{E} ||w^{t_a} - w^*||^2 + 3 \tilde{\eta}^2 \frac{N/M - 1}{N - 1} \sigma_*^2\nonumber  \\
    &~~~+ (6 L \tilde{\eta}^2 (1 + \frac{N/M - 1}{N - 1}) - 2 \tilde{\eta}) \mathbb{E}(F(w^{t_a}) - F(w^*))\nonumber  \\
    &~~~+ (\frac{2}{\mu_F} + 3\tilde{\eta})\tilde{\eta} \mathbb{E}(3e \tilde{\eta}^2L 2^{K+1} \nonumber \\
    &~~~\frac{4L(1 + 2K)(F(w^{t_a}) - F(w^*)) + 2\sigma_*^2 + \dot{\delta}}{K^2} + \dot{\delta}).
\end{align}

Let $\dot{\delta}_1 := \frac{8 \dot{\delta}}{\mu_F}$, $\dot{\delta}_2 := 3 \sigma_*^2 \frac{N/M - 1}{N - 1}$ and $\dot{\delta}_3 := 48eL2^K \frac{2 \sigma_*^2 + \dot{\delta}}{K^2}$, we have:
\begin{align}
    \mathbb{E} \Delta^{t_a + K} \leq (1 - \frac{\tilde{\eta} \mu_F}{2}) \mathbb{E} \Delta^{t_a} &- \tilde{\eta} \mathbb{E}(F(w^{t_a}) - F(w^*)) \nonumber \\
    &~~~+ \sum_{j=1}^3 \dot{\delta}_j \tilde{\eta}^j \\
    \Rightarrow F(w^{t_a}) - F(w^*) &\leq \frac{1}{\tilde{\eta}} \mathbb{E} ((1 - \tilde{\eta})\Delta^{t_a} - \Delta^{t_a + K}) \nonumber \\
    &~~~+ \sum_{j=1}^3 \dot{\delta}_j \tilde{\eta}^{j-1}.
\end{align}

With convexity of $F$, taking $\alpha_t = (1 - \tilde{\eta} \mu_F)^{-t-1}, \bar{w} := \frac{\sum_{t=0}^{(R-K)/K} \alpha_t w^{tK}}{A_R}$ and $A_R = \sum_{t=0}^{(R-K) / K} \alpha_t = \frac{2 \alpha_{R-K}(1-(1 - \frac{\tilde{\eta}}{2})^{R/K})}{\tilde{\eta} \mu_F}$ multipied on both side, we have:
\begin{align}
    F(\bar{w}) &- F(w^*) \leq \frac{\sum_{t=0}^{(R-K)/K} \alpha_t}{A_R} (F(w^{tK}) - F(w^*)) \\
    &\leq \frac{\sum_{t=0}^{(R-K)/K}}{\tilde{\mu}A_R} \mathbb{E}((1 - \frac{\tilde{\eta} \mu_F}{2}) \alpha_t \Delta^{tK} - \alpha_t \Delta^{(t+1)K}) \nonumber \\
    &~~~+ \sum_{j=1}^{3} \dot{\delta}_j \tilde{\eta}^{j-1} \\
    &= \frac{1}{\tilde{\eta} A_R} \mathbb{E}((1 - \frac{\tilde{\eta} \mu_F}{2}) \Delta^0 - \alpha_{R-K} \Delta^T) \nonumber \\
    &~~~+ \sum_{j=1}^{3} \dot{\delta}_j \tilde{\eta}^{j-1} \\
    &= \frac{\mu_F}{2 \alpha_{R-K} (1 - (1 -\tilde{\eta} \mu_F / 2) ^ {R/K})} \Delta^0 \nonumber \\
    &~~~- \frac{\alpha_{R - K}}{\tilde{\eta} A_R} \Delta^T + \sum_{j=1}^{3} \dot{\delta}_j \tilde{\eta}^{j-1}.
\end{align}

By setting $\tilde{\eta} R \geq \frac{2}{\mu_F}$ and the fact $\frac{\alpha_{R-K}}{\tilde{\eta} A_R} \geq A_R \geq \frac{\alpha_{R-K}}{\tilde{\eta} \mu_F}$, we have:
\begin{align}
    F(\bar{w}) - F(w^*) &\leq \mu_F e^{-\tilde{\eta}\mu_F \frac{R}{2}} \Delta^0 - \frac{\mu_F}{2} \Delta^T + \sum_{j=1}^{3} \dot{\delta}_j \tilde{\eta}^{j-1} \\
    &\leq \mu_F e^{-\tilde{\eta}\mu_F \frac{R}{2}} \Delta^0 + \sum_{j=1}^{3} \dot{\delta}_j \tilde{\eta}^{j-1}.
\end{align}

Finally, the following equation can be derived:
\begin{align}
    &~~~\mathcal{O}(F(\bar{w}) - F(w^*)) \nonumber \\
    &= \mathcal{O}(\mu_F e^{-\tilde{\eta} \mu_F \frac{R}{2} \Delta^0}) + \mathcal{O}(\frac{\kappa_S^2 + \sigma_F^2}{\mu_F}) \nonumber \\
    &~~~+ \mathcal{O}(\frac{N/M - 1}{N \mu_F R} \sigma_*^2) + \mathcal{O}(\frac{2^K L (\sigma_*^2 + \kappa_S^2 + \sigma_F^2)}{K^2 \mu_F^2 R^2}).
\end{align}

Therefore, the Theorem~\ref{theorem:theorem_1} is proved.
\end{proof}

\section*{Proof of Theorem~\ref{theorem:theorem_2}}
\begin{proof}
\begin{align}
    &~~~\mathbb{E}(F(w^{t_a + K}) - F(w^{t_a})) \nonumber \\
    &\leq \mathbb{E} \langle \nabla F(w^{t_a}), w^{t_a + K} - w^{t_a} \rangle + \frac{L}{2} \mathbb{E}||w^{t_a + K} - w^{t_a}||^2 \\
    &= -\tilde{\eta} \mathbb{E} ||\nabla F(w^{t_a})||^2 + \frac{L \tilde{\eta}^2}{2} \mathbb{E}||\tilde{\eta}^{-1} \sum_{k=0}^{K - 1} g_i^{t_a + k}||^2 \nonumber \\
    &~~~- \tilde{\eta} \mathbb{E} \langle F(w^{t_a}), \tilde{\eta}^{-1} \sum_{k=0}^{K-1} g_i^{t_a + k} - \nabla F(w^{t_a}) \rangle \\
    &\leq -\tilde{\eta} \mathbb{E} ||\nabla F(w^{t_a})||^2 + \frac{L \tilde{\eta}^2}{2} \mathbb{E}||\tilde{\eta}^{-1} \sum_{k=0}^{K - 1} g_i^{t_a + k}||^2 \nonumber \\
    &~~~+ \frac{\tilde{\eta}}{2} \mathbb{E} ||\tilde{\eta}^{-1} \sum_{k=0}^{K-1} g_i^{t_a + k} - \nabla F(w^{t_a})||^2 \\
    &\leq -\tilde{\eta} \mathbb{E} ||\nabla F(w^{t_a})||^2 + \frac{L \tilde{\eta}^2}{2} \frac{1}{NK} \sum_{i=0, k=0}^{N, K-1} \mathbb{E}||\tilde{\eta}^{-1} g_i^{t_a + k}||^2 \nonumber \\
    &~~~+ \frac{\tilde{\eta}}{2} \frac{1}{NK} \sum_{i=0, k=0}^{N, K-1} \mathbb{E}||\tilde{\eta}^{-1} g_i^{t_a + k} - \nabla F(w^{t_a})||^2 \\
    &\leq \frac{(3 L \tilde{\eta} - 1)\tilde{\eta}}{2} \mathbb{E} ||\nabla F(w^{t_a})||^2 \nonumber \\
    &~~~+ \frac{3L\tilde{\eta}^2}{2} \frac{N/M - 1}{N(N-1)} \sum_{i=1}^{N} \mathbb{E}||\nabla F(w^{t_a}) - \nabla F(w^{t_a})||^2 \nonumber \\
    &~~~+ \frac{\tilde{\eta} (1 + 3L\tilde{\eta})}{2} \frac{1}{NK} \sum_{i=0}^{N} \mathbb{E}||\tilde{\eta}^{-1} g_i^{t_a + k} - \nabla F(w^{t_a})||^2.
\end{align}

Taking Lemma~\ref{lemma:bounded_local_approximation} into the above equation, we have:
\begin{align}
     &~~~\mathbb{E}(F(w^{t_a + K}) - F(w^{t_a})) \nonumber \\
     &\leq \frac{(3 L \tilde{\eta} - 1)\tilde{\eta}}{2} \mathbb{E} ||\nabla F(w^{t_a})||^2 \nonumber \\
     &~~~+ \frac{3L\tilde{\eta}^2}{2} \frac{N/M - 1}{N(N-1)} \sum_{i=1}^{N} \mathbb{E}||\nabla F(w^{t_a}) - \nabla F(w^{t_a})||^2\nonumber  \\
     &~~~+ \frac{\tilde{\eta}(1 + 3L\tilde{\eta})}{2} \frac{1}{N} \sum_{i=1}^N \mathbb{E}(3 \eta^2 Le2^{K + 1}(2(1 + 2K)\nonumber  \\
     &~~~~~~||\nabla F(w^{t_a}) - \nabla F(w^{t_a})||^2 + 2(1 + 2K)||\nabla F(w^{t_a})||^2 \nonumber \\
     &~~~~~~+ \dot{\delta}) + \dot{\delta}) \\
     &\leq \frac{(3 L \tilde{\eta} - 1)\tilde{\eta}}{2} \mathbb{E} ||\nabla F(w^{t_a})||^2 \nonumber \\
     &~~~+ \frac{3L\tilde{\eta}^2}{2} \frac{N/M - 1}{N(N-1)} \sum_{i=1}^{N} \mathbb{E}||\nabla F(w^{t_a}) - \nabla F(w^{t_a})||^2 \nonumber \\
     &~~~+ \frac{\tilde{\eta}(1 + 3L\tilde{\eta})}{2} \frac{1}{N} \sum_{i=1}^N 3 \eta^2 Le2^{K + 2}(1 + 2K) \nonumber \\
     &~~~~~~\mathbb{E}(||\nabla F(w^{t_a}) - \nabla F(w^{t_a})||^2 + ||\nabla F(w^{t_a})||^2) \nonumber \\
     &~~~~~~+ \frac{\tilde{\eta}(1 + 3L\tilde{\eta})(3 \tilde{\eta}^2 Le 2^{K + 1}/K^2 + 1)}{2} \dot{\delta} \\
     &\leq (\frac{\tilde{\eta}(3L\tilde{\eta} - 1)}{2} + \frac{3L \tilde{\eta}}{2} \frac{N / M - 1}{N - 1} \nonumber \\
     &~~~+ (1 + \lambda_F^2)\frac{\tilde{\eta}(1 + 3L\tilde{\eta})}{2K} 3\tilde{\eta}^2 Le 2^{K + 2}(1 / K + 2))\mathbb{E}||\nabla F(w^{t_a})||^2 \nonumber \\
     &~~~+ \frac{3Le2^{K + 2}(1/K + 2)\delta_F^2 \tilde{\eta}^3 (1 + 3L\tilde{\eta})}{2K} + \frac{3L\tilde{\eta}^2 \delta_F^2}{2} \frac{N / M - 1}{N - 1} \nonumber \\
     &~~~+ \frac{\tilde{\eta} (1 + 3L\tilde{\eta}) (3 \tilde{\eta}^2 Le2^{K + 1}/ K^2 + 1)}{2}\dot{\delta},
\end{align}
with $\tilde{\eta} \leq \frac{1}{L}$ and $\eta \leq \hat{\eta}_n := \frac{1}{(9LK + 27e2^{K+4})(1 + \lambda_F^2)}$, we have:
\begin{align}
     &~~~\mathbb{E}(F(w^{t_a + K}) - F(w^{t_a})) \nonumber \\
     &\leq -\frac{\tilde{\eta}}{3} \mathbb{E} ||\nabla F(w^{t_a})||^2 + (\frac{9e2^{K+3}L\delta_F^2}{K} + \frac{3e2^{K+2}L\dot{\delta}}{K^2})\tilde{\eta}^3 \nonumber \\
     &~~~+ \frac{3L\delta_F^2}{2} \frac{N / M - 1}{N - 1} \tilde{\eta}^2 + 2\dot{\delta} \tilde{\eta}.
\end{align}

Thus, let $C_0 = \frac{2\dot{\delta}}{K}$, $C_1 = \frac{3L \delta_F^2}{2K} \frac{N / M - 1}{N - 1}$ and $C_2 = \frac{9e2^{K+3}L\delta_F^2}{K^2} + \frac{3e2^{K+2}L\dot{\delta}}{K^3}$, we have:
\begin{align}
    \frac{1}{3R} \sum_{t_a \in R_a} \mathbb{E} ||\nabla F(w^{t_a})||^2 &\leq \frac{1}{\tilde{\eta}^R} \mathbb{E}(F(w^0) - F(w^R)) + \sum_{j=0}^2 C_j \tilde{\eta}^j.
\end{align}

Let $\Delta^* = \mathbb{E}(F(w^0) - F(w^*))$, the above inequation can be derived to:
\begin{align}
    \frac{1}{3R} \sum_{t_a \in R_a} \mathbb{E} ||\nabla F(w^{t_a})||^2 &\leq \frac{1}{\tilde{\eta}^R} \Delta^* + \sum_{j=0}^2 C_j \tilde{\eta}^j.
\end{align}

Consequently, by uniformly sampling a $t^* \in R_a$, there are in total 2 cases that need to be discussed separately:
\begin{enumerate}
    \item When $\frac{\delta^*}{RC_1} \leq \hat{\eta}_n^2$ or $\frac{\delta^*}{C_2} \leq \hat{\eta}_n^3$, by choosing a proper $\tilde{\eta} = \min\{\frac{{\delta^*}^{\frac{1}{2}}}{C_{1}^{\frac{1}{2}}R^{\frac{1}{2}}}, \frac{{\Delta^*}^{\frac{1}{3}}}{C_{2}^{\frac{1}{3}}R^{\frac{1}{3}}}\}$, the following bound can be derived:
    \begin{align}
        \mathcal{O}(\mathbb{E}||\nabla F(w^{t^*})||^2) = K(C_0 +  \frac{{\Delta^*}^{\frac{1}{2}}C_{1}^{\frac{1}{2}}}{R^{\frac{1}{2}}} + \frac{{\Delta^*}^{\frac{2}{3}}C_{2}^{\frac{1}{3}}}{R^{\frac{2}{3}}}).
    \end{align}
    \item When $\frac{\Delta^*}{RC_{1}} \ge \hat{\eta}_{n}^{2}$ or $\frac{\Delta^*}{C_{2}} \ge \hat{\eta}_{n}^{3}$, by choosing a proper $\tilde{\eta} = \hat{\eta}_{n}$, the following bound can be derived:
    \begin{align}
        \mathcal{O}(\mathbb{E}||\nabla F(w^{t^*})||^2) = K(\frac{1}{\tilde{\eta}R}\Delta^* + C_{0} + \frac{{\Delta^*}^{\frac{1}{2}}C_{1}^{\frac{1}{2}}}{R^{\frac{1}{2}}} + \frac{{\Delta^*}^{\frac{2}{3}}C_{2}^{\frac{1}{3}}}{R^{\frac{2}{3}}}).
    \end{align}
\end{enumerate}

Finally, a general bound can be derived as follows, and the Theorem~\ref{theo_2} is therefore proved:
\begin{align}
     &~~~\mathcal{O}(\mathbb{E}||\nabla F(w^{t^*})||^2) \nonumber \\
     &= \mathcal{O}(\frac{\Delta^*}{\tilde{\eta}R}) + \mathcal{O}(\dot{\delta}) + \mathcal{O}(\frac{{\Delta^*}^{\frac{1}{2}}L^{\frac{1}{2}}\delta_{F} (\frac{N/M-1}{N})^{\frac{1}{2}}}{K^{\frac{1}{2}}R^{\frac{1}{2}}}) \nonumber \\
     &~~~+ \mathcal{O}(\frac{{\Delta^*}^{\frac{2}{3}}2^{\frac{K}{3}}L^{\frac{1}{3}}}{KR^{\frac{2}{3}}} (L\delta_{F}^{2}K + L\dot{\delta})^{\frac{1}{3}}).
\end{align}
\end{proof}
}

\end{document}